\documentclass[journal]{IEEEtran}

\usepackage{amsmath,amssymb,amsfonts,bm}
\usepackage{graphicx}
\usepackage{booktabs,array,multirow}
\usepackage[table]{xcolor}\usepackage{pifont}
\usepackage{textcomp,microtype,url}
\usepackage[caption=false,font=footnotesize]{subfig}
\usepackage{capt-of}
\usepackage[nocompress]{cite}
\usepackage{placeins}
\usepackage{dblfloatfix}
\usepackage{needspace}
\usepackage{float}
\usepackage{algorithm}
\usepackage{algpseudocode}
\algrenewcommand\algorithmicrequire{\textbf{Input:}}
\algrenewcommand\algorithmicensure{\textbf{Output:}}
\usepackage{hyperref}
\hypersetup{
    colorlinks=true,
    linkcolor=blue,
    citecolor=blue,
    urlcolor=blue
}

\providecommand{\captionsetup}[2][]{}
\newcommand{\appqualroot}{figures/Comparison}

\newcommand{\appquallabelfont}{\fontsize{6.3pt}{6.7pt}\selectfont}
\newcommand{\appqualcaptionfont}{\fontsize{8.2pt}{9.0pt}\selectfont}

\newcommand{\appqualportraitscale}{1.0}
\newcommand{\appqualportraitcell}[2]{%
    \begin{minipage}[t]{0.1460\linewidth}
        \centering
        \includegraphics[width=\appqualportraitscale\linewidth,keepaspectratio]{#1}\\[0.30mm]
        {\appquallabelfont #2}
    \end{minipage}%
}

\newcommand{\appqualhknasportrait}[2]{%
    \IfFileExists{\appqualroot/#1/HKNAS-3D.pdf}{%
        \appqualportraitcell{\appqualroot/#1/HKNAS-3D.pdf}{#2}%
    }{%
        \appqualportraitcell{\appqualroot/#1/HKNAS_3D.pdf}{#2}%
    }%
}

\newcommand{\appqualwideboxheight}{6.2mm}
\newcommand{\appqualwidecell}[2]{%
    \begin{minipage}[t]{0.154\linewidth}
        \centering
        \begin{minipage}[c][\appqualwideboxheight][c]{\linewidth}
            \centering
            \includegraphics[width=\linewidth,height=5.5mm,keepaspectratio]{#1}%
        \end{minipage}\\[-0.55mm]
        {\appquallabelfont #2}
    \end{minipage}%
}

\newcommand{\appqualwidecellrot}[2]{%
    \begin{minipage}[t]{0.154\linewidth}
        \centering
        \begin{minipage}[c][\appqualwideboxheight][c]{\linewidth}
            \centering
            \rotatebox{90}{\includegraphics[height=\linewidth,keepaspectratio]{#1}}%
        \end{minipage}\\[-0.55mm]
        {\appquallabelfont #2}
    \end{minipage}%
}

\newcommand{\appqualhknaswide}[2]{%
    \IfFileExists{\appqualroot/#1/HKNAS_3D.pdf}{%
        \appqualwidecell{\appqualroot/#1/HKNAS_3D.pdf}{#2}%
    }{%
        \appqualwidecellrot{\appqualroot/#1/HKNAS-3D.pdf}{#2}%
    }%
}

\newcommand{\appqualportraitgrid}[1]{%
    \begingroup\setlength{\tabcolsep}{0pt}%
    \noindent\begin{tabular*}{\linewidth}{@{\extracolsep{\fill}}cccccc@{}}
        \appqualportraitcell{\appqualroot/#1/GT.pdf}{(a)} &
        \appqualportraitcell{\appqualroot/#1/CVSSN.pdf}{(b)} &
        \appqualportraitcell{\appqualroot/#1/SSRN.pdf}{(c)} &
        \appqualhknasportrait{#1}{(d)} &
        \appqualportraitcell{\appqualroot/#1/SSFTT.pdf}{(e)} &
        \appqualportraitcell{\appqualroot/#1/GAHT.pdf}{(f)}
    \end{tabular*}\\[0.65mm]
    \noindent\begin{tabular*}{\linewidth}{@{\extracolsep{\fill}}cccccc@{}}
        \appqualportraitcell{\appqualroot/#1/CMT.pdf}{(g)} &
        \appqualportraitcell{\appqualroot/#1/MambaHSI.pdf}{(h)} &
        \appqualportraitcell{\appqualroot/#1/3DSS-Mamba.pdf}{(i)} &
        \appqualportraitcell{\appqualroot/#1/MambaLG.pdf}{(j)} &
        \appqualportraitcell{\appqualroot/#1/SliMamba.pdf}{(k)} &
        \appqualportraitcell{\appqualroot/#1/Compact48.pdf}{(l)}
    \end{tabular*}%
    \endgroup
}

\newcommand{\appqualhoustonboxheight}{17.2mm}
\newcommand{\appqualhoustonlabelfont}{\fontsize{8pt}{8.6pt}\selectfont}
\newcommand{\appqualhoustoncell}[2]{%
    \begin{minipage}[t]{0.485\linewidth}
        \centering
        \begin{minipage}[c][\appqualhoustonboxheight][c]{\linewidth}
            \centering
            \includegraphics[width=\linewidth,height=15.8mm,keepaspectratio]{#1}%
        \end{minipage}\\[0.30mm]
        {\appqualhoustonlabelfont #2}
    \end{minipage}%
}

\newcommand{\appqualhoustoncellrot}[2]{%
    \begin{minipage}[t]{0.485\linewidth}
        \centering
        \begin{minipage}[c][\appqualhoustonboxheight][c]{\linewidth}
            \centering
            \rotatebox{90}{\includegraphics[height=.92\linewidth,keepaspectratio]{#1}}%
        \end{minipage}\\[0.30mm]
        {\appqualhoustonlabelfont #2}
    \end{minipage}%
}

\newcommand{\appqualhknashouston}[2]{%
    \IfFileExists{\appqualroot/#1/HKNAS_3D.pdf}{%
        \appqualhoustoncell{\appqualroot/#1/HKNAS_3D.pdf}{#2}%
    }{%
        \appqualhoustoncellrot{\appqualroot/#1/HKNAS-3D.pdf}{#2}%
    }%
}

\newcommand{\appqualhoustongrid}[1]{%
    \begingroup\setlength{\tabcolsep}{0pt}%
    \noindent\begin{tabular*}{\linewidth}{@{\extracolsep{\fill}}cc@{}}
        \appqualhoustoncell{\appqualroot/#1/GT.pdf}{(a)} &
        \appqualhoustoncell{\appqualroot/#1/CVSSN.pdf}{(b)}
    \end{tabular*}\\[0.45mm]
    \noindent\begin{tabular*}{\linewidth}{@{\extracolsep{\fill}}cc@{}}
        \appqualhoustoncell{\appqualroot/#1/SSRN.pdf}{(c)} &
        \appqualhknashouston{#1}{(d)}
    \end{tabular*}\\[0.45mm]
    \noindent\begin{tabular*}{\linewidth}{@{\extracolsep{\fill}}cc@{}}
        \appqualhoustoncell{\appqualroot/#1/SSFTT.pdf}{(e)} &
        \appqualhoustoncell{\appqualroot/#1/GAHT.pdf}{(f)}
    \end{tabular*}\\[0.45mm]
    \noindent\begin{tabular*}{\linewidth}{@{\extracolsep{\fill}}cc@{}}
        \appqualhoustoncell{\appqualroot/#1/CMT.pdf}{(g)} &
        \appqualhoustoncell{\appqualroot/#1/MambaHSI.pdf}{(h)}
    \end{tabular*}\\[0.45mm]
    \noindent\begin{tabular*}{\linewidth}{@{\extracolsep{\fill}}cc@{}}
        \appqualhoustoncell{\appqualroot/#1/3DSS-Mamba.pdf}{(i)} &
        \appqualhoustoncell{\appqualroot/#1/MambaLG.pdf}{(j)}
    \end{tabular*}\\[0.45mm]
    \noindent\begin{tabular*}{\linewidth}{@{\extracolsep{\fill}}cc@{}}
        \appqualhoustoncell{\appqualroot/#1/SliMamba.pdf}{(k)} &
        \appqualhoustoncell{\appqualroot/#1/Compact48.pdf}{(l)}
    \end{tabular*}%
    \endgroup
}

\newcommand{\appqualhoustonfigure}[3]{%
    \noindent\begin{minipage}{\linewidth}
        \centering
        \appqualhoustongrid{#1}
        \vspace{-0.35em}
        {\appqualcaptionfont\captionof{figure}{#2 \appqualcaptionmethods}\label{#3}}
    \end{minipage}%
}

\newcommand{\appqualcaptionmethods}{%
(a) Ground truth. (b) CVSSN. (c) SSRN. (d) HKNAS3D. (e) SSFTT. (f) GAHT.
(g) CMT. (h) MambaHSI. (i) 3DSS-Mamba. (j) MambaLG. (k) SliMamba. (l) Compact48.%
}

\newcommand{\appqualportraitfigure}[3]{%
    \noindent\begin{minipage}{\linewidth}
        \centering
        \appqualportraitgrid{#1}
        \vspace{-0.60em}
        {\appqualcaptionfont\captionof{figure}{#2 \appqualcaptionmethods}\label{#3}}
    \end{minipage}%
}

\newcommand{\appqualhanonecell}[2]{%
    \begin{minipage}[t]{0.079\linewidth}%
        \centering
        \includegraphics[width=\linewidth,keepaspectratio]{#1}\\[-0.55mm]
        {\appquallabelfont #2}%
    \end{minipage}%
}
\newcommand{\appqualhanonegrid}[1]{%
    \begingroup\setlength{\tabcolsep}{0pt}%
    \noindent\begin{tabular*}{\linewidth}{@{\extracolsep{\fill}}*{12}{c}@{}}
        \appqualhanonecell{\appqualroot/#1/GT.pdf}{(a)} &
        \appqualhanonecell{\appqualroot/#1/CVSSN.pdf}{(b)} &
        \appqualhanonecell{\appqualroot/#1/SSRN.pdf}{(c)} &
        \appqualhanonecell{\appqualroot/#1/HKNAS-3D.pdf}{(d)} &
        \appqualhanonecell{\appqualroot/#1/SSFTT.pdf}{(e)} &
        \appqualhanonecell{\appqualroot/#1/GAHT.pdf}{(f)} &
        \appqualhanonecell{\appqualroot/#1/CMT.pdf}{(g)} &
        \appqualhanonecell{\appqualroot/#1/MambaHSI.pdf}{(h)} &
        \appqualhanonecell{\appqualroot/#1/3DSS-Mamba.pdf}{(i)} &
        \appqualhanonecell{\appqualroot/#1/MambaLG.pdf}{(j)} &
        \appqualhanonecell{\appqualroot/#1/SliMamba.pdf}{(k)} &
        \appqualhanonecell{\appqualroot/#1/Compact48.pdf}{(l)}
    \end{tabular*}%
    \endgroup
}
\newcommand{\appqualhanonefigure}[3]{%
    \noindent\begin{minipage}{\linewidth}
        \centering
        \appqualhanonegrid{#1}%
        \par\vspace{0.40mm}%
        {\appqualcaptionfont\captionof{figure}{#2 \appqualcaptionmethods}\label{#3}}%
    \end{minipage}%
}

\title{KGLP-Net: Kalman-Reliable Global--Local Learning With Group Fisher Pruning for Hyperspectral Image Classification}

\author{Wenshuai~Deng,
Wen~Xie,~\IEEEmembership{Member,~IEEE},
Abdel~Hamid~Soliman, and~Qianzheng~Zhao%
\thanks{This work has been submitted to the IEEE for possible publication. Copyright may be transferred without notice, after which this version may no longer be accessible.}%
\thanks{This work was supported in part by the National
Natural Science Foundation of China under Grants 61901365
and 62071379, in part by the Natural Science Basic Research
Plan in Shaanxi Province of China under Grant 2026JC-YBMS-0799,
in part by the Scientific Research Program Project of the
Youth Innovation Team of Shaanxi Provincial Department of
Education under Grant 25JP175, and in part by the Youth
Innovation Team of Shaanxi Universities.
\emph{(Corresponding author: Wen Xie.)}}%
\thanks{Wenshuai Deng, Wen Xie, and Qianzheng Zhao are
with the School of Communications and Information
Engineering, Xi'an University of Posts and Telecommunications,
Xi'an 710121, China (e-mail: shuaidyb@stu.xupt.edu.cn;
xiewen@xupt.edu.cn; zqz@stu.xupt.edu.cn).}%
\thanks{Abdel Hamid Soliman is with the School of
Digital, Technology, Innovation, and Business,
University of Staffordshire, Stoke-on-Trent ST4 2DE,
U.K. (e-mail: a.soliman@staffs.ac.uk).}%
}

\begin{document}

\maketitle

\begin{abstract}
Accurate hyperspectral image (HSI) classification requires both the effective integration of broad spatial context and fine-grained spectral--spatial information surrounding target pixels. However, the reliability of these complementary representations may vary considerably across heterogeneous scenes, and uncertainty estimates obtained during feature fusion are rarely exploited to guide subsequent model compression. To address these challenges, we propose Kalman-Reliable Global--Local Learning with Group Fisher Pruning (KGLP-Net), a unified framework that couples uncertainty-aware feature fusion with reliability-guided structured pruning. Specifically, an asymmetric dual-view architecture employs a Global Mamba Encoder to capture long-range spatial dependencies and a Local CNN Encoder to extract fine-grained spectral--spatial features from target neighborhoods. To adaptively integrate these complementary representations, Kalman-Reliable Fusion (KRF) estimates their relative uncertainties in feature space and performs a Kalman-inspired posterior update, treating global features as a contextual prior and local features as an observation. Building upon the resulting posterior uncertainty, Kalman-Reliable Group Fisher Pruning (KRGF) quantifies group-level reliability through normalized posterior precision and integrates it with Group Fisher importance to identify and prune Kalman-coupled channel groups, enabling structurally consistent model compression. Experiments on four public HSI datasets demonstrate that the resulting Compact48 model, with a hidden width of 48, achieves the highest overall accuracy among ten representative competing classifiers. Compared with the unpruned Dense64 model, Compact48 reduces the number of parameters, FLOPs, and inference time by 38.5\%, 34.8\%, and 37.6\% on average, respectively, while incurring only a 0.03-percentage-point decrease in mean overall accuracy.
\end{abstract}


\section{Introduction}

Hyperspectral images (HSIs) record densely sampled spectral responses together
with spatial context, supporting fine-grained discrimination of land-cover
materials. Pixel-wise classification nevertheless remains challenging because
spectra from the same category can vary with illumination, material composition,
and acquisition conditions, whereas different categories may exhibit similar
spectral signatures. Mixed pixels and transitions between neighboring classes
further complicate discrimination near land-cover boundaries~\cite{ghamisi2018new,he2018recent,li2019deep}. These properties make the
spatial neighborhood important for interpreting a target spectrum, but the
neighborhood itself may contain multiple materials. Consequently, accurate HSI
classification requires representations that capture informative contextual
relationships without obscuring the spectral--spatial evidence surrounding
the target pixel.

Deep HSI classifiers have progressively extended their ability to model such
relationships. Convolutional architectures learn local spectral--spatial
patterns, including through three-dimensional filtering~\cite{chen2014deep,benhamida2018deep,zhong2018ssrn,roy2020hybridsn}.
Transformer-based approaches capture wider dependencies through self-attention~\cite{vaswani2017attention,hong2022spectralformer,sun2022ssftt,mei2022gaht},
while structured state-space models and Mamba offer efficient alternatives for
long-range representation learning~\cite{gu2022efficient,gu2024mamba,zhu2024visionmamba,liu2024vmamba,
li2024mambahsi}. Global--local and dual-branch HSI architectures further
combine broad context with finer target-neighborhood information~\cite{qi2023global,zhou2024dctn,meng2025glmgt,pan2025mamba,
sheng2025dualmamba}. These developments establish the value of complementary
spatial supports. For pixel-centered classification, however, combining
complementary representations also requires determining how strongly each
source should influence the prediction.

\begin{figure*}[!t]
    \centering
    \includegraphics[width=\linewidth]{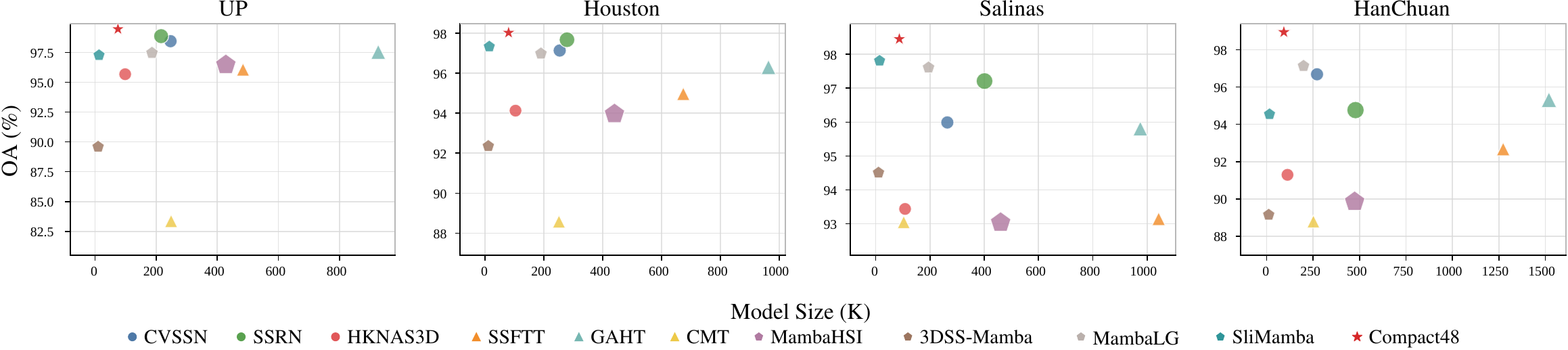}
    \caption{Accuracy--compactness comparison of representative HSI classifiers on
    UP, Houston, Salinas, and HanChuan. The horizontal and vertical axes denote
    parameter count (K) and OA (\%), respectively; marker size represents FLOPs (M).}
    \label{fig:accuracy_compactness}
\end{figure*}

The distinction arises from the spatial--spectral heterogeneity of HSI scenes.
Within homogeneous regions, a broad neighborhood can reinforce coherent
land-cover patterns and help discriminate spectrally similar materials. Near
class boundaries, within fragmented parcels, or around small objects, the same
neighborhood may include responses from other categories. A smaller
center-aligned view better preserves local evidence, yet it may lack the
context needed to resolve an ambiguous target spectrum. Thus, the relative
usefulness of global and local information depends on their spatial support
and the observed features, rather than on a fixed preference for one branch.
Common operations such as addition, concatenation, attention, and gating
combine feature responses, but their fusion weights do not necessarily
represent the uncertainty associated with each source. This motivates
explicitly estimating the relative feature-space uncertainty of global and
local representations when integrating them for HSI classification.

The uncertainty estimated during fusion also raises a related question for
lightweight model design: can it inform which hidden structures should be
retained? HSI classifiers process high-dimensional spectral measurements over
many target pixels, making parameter count and computational cost important.
Existing lightweight architectures reduce complexity through compact
convolutions, spectral--spatial redesign, and efficient attention~\cite{jia2021lightweight,zhou2023adaptive,li2024cloln,zhao2024gscvit}.
Structured pruning provides another route by removing channels or filters
while preserving regular computation~\cite{li2017pruning,he2017channelpruning,luo2017thinet,liu2017slimming}.
Task-aware and dependency-aware pruning criteria, including Taylor-based and
Group Fisher importance, estimate the classification sensitivity of structural
units and account for coupled dependencies~\cite{molchanov2019importance,liu2021group,fang2023depgraph}. Such importance
is distinct from the relative reliability estimated during feature fusion:
the former describes a group's contribution to the task objective, whereas
the latter reflects the relative precision associated with its fused
information. Their combination offers a way to carry reliability information
from representation learning into structural selection, rather than treating
fusion and compression as unrelated stages.

Motivated by these considerations, we propose \textbf{Kalman-Reliable
Global--Local Learning with Group Fisher Pruning (KGLP-Net)}, a framework
linking uncertainty-aware fusion and reliability-guided structured compression.
An asymmetric dual-view architecture processes a larger patch with a Global
Mamba Encoder to capture contextual dependencies and a smaller center-aligned
patch with a Local CNN Encoder to preserve target-neighborhood cues.
Kalman-Reliable Fusion (KRF) treats the global and local representations as a
contextual prior and an observation, respectively. It estimates their
relative uncertainties from cross-view interactions and performs a
Kalman-inspired feature update, yielding a fused representation and an
associated posterior uncertainty estimate. Kalman-Reliable Group Fisher
Pruning (KRGF) then derives group-level posterior reliability from this
uncertainty and combines it with Group Fisher task sensitivity to select
Kalman-coupled global, local, and fused channels. In this way, an uncertainty
signal produced during global--local learning also influences the structures
retained in the compact network. Fig.~\ref{fig:accuracy_compactness}
illustrates the accuracy--compactness trade-off of Compact48, the model
obtained by KRGF compression of Dense64 from hidden width 64 to 48,
relative to representative HSI classifiers. The main contributions are summarized as
follows:

\begin{enumerate}
    \item \textbf{Uncertainty-Aware Global--Local Representation Learning.}
    We develop an asymmetric dual-view framework for pixel-centered HSI
    classification, integrating a Global Mamba Encoder for long-range contextual modeling and a Local CNN Encoder for fine-grained spectral--spatial feature extraction. To account for the varying reliability of these complementary representations, we propose KRF, which estimates their
    relative feature-space uncertainties and performs a Kalman-inspired
    posterior update. This formulation enables uncertainty-aware feature integration while providing posterior uncertainty estimates for subsequent reliability assessment.

    \item \textbf{Reliability-Guided Structured Pruning.}
    We propose KRGF, which explicitly connects fusion-derived posterior reliability with structured model compression. Specifically, we establish Kalman-coupled channel groups to preserve structural dependencies among the global, local, and fused representations. By combining
    fusion-derived posterior reliability
    with Group Fisher task sensitivity, KRGF evaluates channel groups in terms of both feature reliability and task sensitivity, enabling reliability-guided group selection and structurally consistent reconstruction of compact models.

    \item \textbf{Comprehensive Evaluation of Accuracy and Efficiency.}
    Extensive experiments on four public HSI datasets validate the effectiveness of the proposed global--local architecture, KRF, and KRGF. Ablation studies further demonstrate that posterior reliability provides complementary information to Fisher-based importance for channel selection under aggressive compression. The resulting Compact48 model
    achieves the highest OA among ten representative competing classifiers
    across all four datasets, while reducing the parameter count, FLOPs, and
    inference time relative to Dense64 with little change in mean OA.
\end{enumerate}

The remainder of this article is organized as follows.
\hyperref[sec:related_work]{Section II} reviews related work,
\hyperref[sec:method]{Section III} describes KGLP-Net,
\hyperref[sec:experiments]{Section IV} presents the
experimental results and analysis, and
\hyperref[sec:conclusion]{Section V} concludes the article.

\begin{figure*}[!t]
    \centering
    \includegraphics[width=0.98\linewidth]{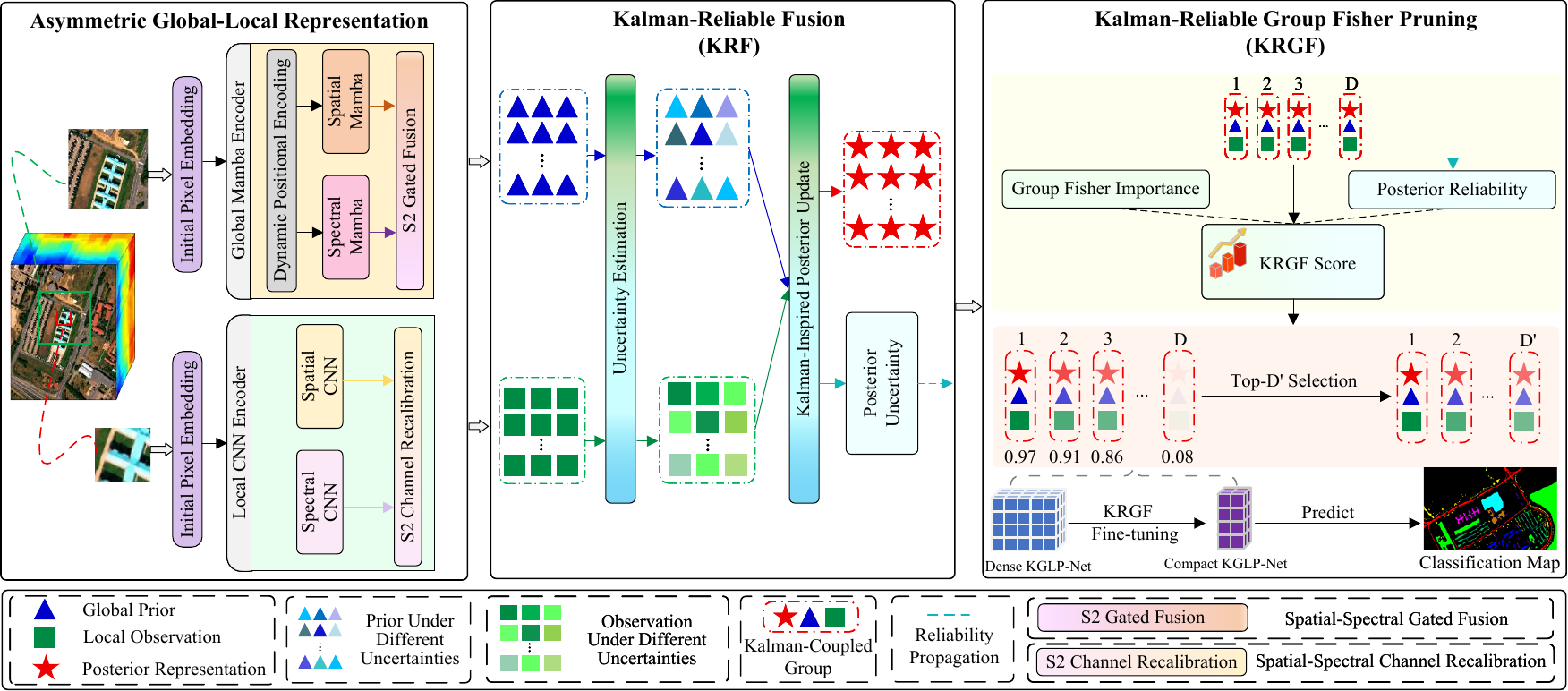}
    \caption{Overview of KGLP-Net. The asymmetric global--local encoders extract
    complementary spectral--spatial representations. Kalman-Reliable Fusion
    (KRF) combines them using learned relative uncertainty estimates, while
    Kalman-Reliable Group Fisher Pruning (KRGF) uses fusion-derived posterior
    reliability and Group Fisher importance to select coupled channel groups
    for compact reconstruction.}
    \label{fig:overview}
\end{figure*}

\section{Related Work}
\label{sec:related_work}

\subsection{Global--Local Hyperspectral Representation Learning}
Deep HSI classification has evolved from local spectral--spatial feature
extraction toward the integration of contextual information across different spatial scales. CNN-based methods, such as SSRN and HybridSN, exploit convolutional
inductive biases to capture local spectral--spatial dependencies~\cite{zhong2018ssrn,roy2020hybridsn}. Center--neighborhood modeling further
recognizes that a target pixel and its surrounding pixels have different
roles in classification~\cite{li2023cvssn}. Transformer-based methods, including
SpectralFormer, SSFTT, and GAHT, extend the modeling range through attention~\cite{hong2022spectralformer,sun2022ssftt,mei2022gaht}, while global--local
architectures such as DCTN and GLMGT integrate complementary representations from different spatial supports~\cite{zhou2024dctn,meng2025glmgt}.

State-space models (SSMs) offer an efficient alternative for contextual representation learning. Structured SSMs and selective Mamba facilitate efficient sequence
modeling~\cite{gu2022efficient,gu2024mamba} and have subsequently been extended to visual representation learning~\cite{zhu2024visionmamba,liu2024vmamba}. HSI-oriented models,
including MambaHSI, 3DSS-Mamba, HyperMamba, and S2Mamba, exploit these mechanisms to capture spectral--spatial dependencies~\cite{li2024mambahsi,he2024_3dssmamba,liu2024hypermamba,wang2025s2mamba}.
MambaLG and DualMamba further combine broader context with local detail~\cite{pan2025mamba,sheng2025dualmamba}. More recent methods explore adaptive spectral--spatial feature interaction through fusion gating in DTCMamba, spectral-guided feature modulation in SMGNet, and channel-wise competitive fusion in PyS$^{2}$CF-Mamba~\cite{li2026dtcmamba,smgnet2026,gao2026pys2cf}. Although these methods demonstrate the benefits of complementary representations, the relative reliability of global and local features may vary across spatially heterogeneous scenes. To address this issue, KGLP-Net introduces Kalman-Reliable Fusion, which explicitly estimates relative feature-space uncertainties through cross-view interactions and uses them to guide adaptive global--local feature integration.

\subsection{Uncertainty and Reliability Modeling}
Uncertainty modeling provides ways to represent ambiguity in learned
predictions or intermediate features. Representative approaches include
Bayesian approximations~\cite{gal2016dropout}, aleatoric and epistemic
uncertainty modeling~\cite{kendall2017uncertainties}, deep ensembles~\cite{lakshminarayanan2017simple}, and evidential learning~\cite{sensoy2018evidential}. Predictive reliability is also studied through
confidence calibration~\cite{guo2017calibration} and uncertainty estimation
under distribution shift~\cite{ovadia2019trust}. While these studies primarily focus on predictive uncertainty and confidence estimation, KGLP-Net considers a different form of reliability, defined as a relative feature-group measure derived from the posterior uncertainty of a learned fusion update.

Uncertainty-aware fusion has also been explored in multi-view learning.
Trusted multi-view classification uses view-specific uncertainty to assess
source trustworthiness and combine evidence~\cite{han2023trusted}, while
trust-based discounting accounts for conflicting information across views~\cite{lu2025navigating}. In remote sensing, uncertainty can guide the
combination of different modalities~\cite{wang2025udfnet}; HSI studies have
also investigated uncertainty for misclassification or out-of-distribution
detection~\cite{yu2024uncertainty} and sample selection or pseudo-label
refinement~\cite{yang2026cabin}. These studies highlight the potential of uncertainty estimation for assessing information reliability. Building on this perspective, KGLP-Net introduces KRF, which performs a Kalman-inspired measurement update in feature space~\cite{kalman1960new} and
provides posterior uncertainty from learned relative uncertainty estimates. The resulting posterior uncertainty is further transformed into group-level reliability scores to guide structured pruning, thereby establishing a connection between uncertainty-aware feature fusion and compact model design.

\begin{figure*}[!t]
    \centering

    \subfloat[Global Mamba Encoder.\label{fig:global_encoder_details}]{%
\begin{minipage}[t]{0.98\linewidth}

        \centering
        \includegraphics[width=\linewidth]
        {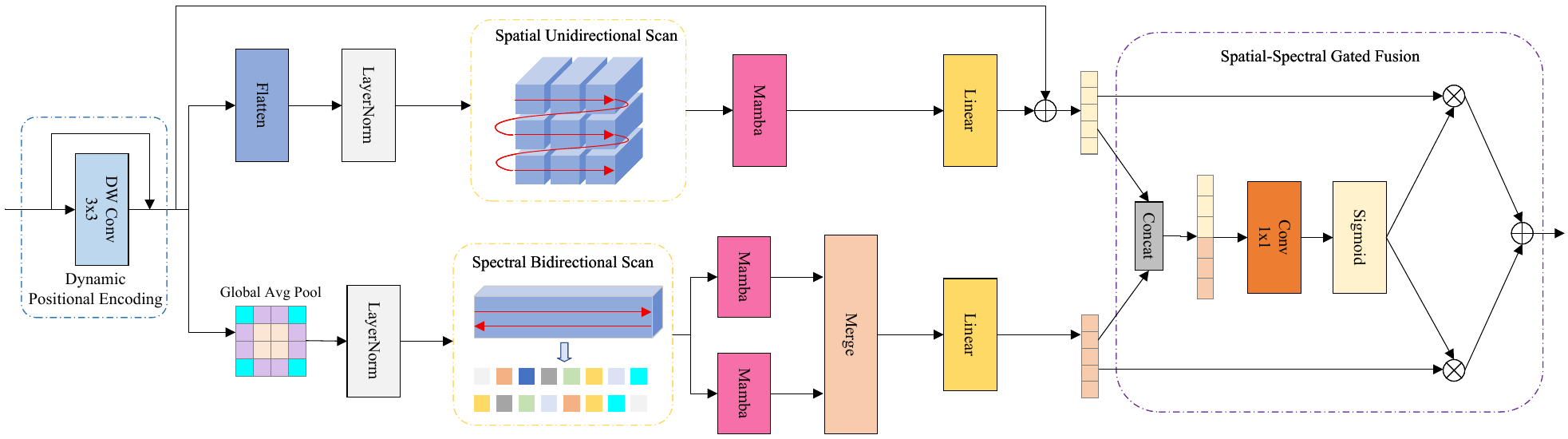}
        \end{minipage}%
}

    \vspace{-0.8mm}

    \subfloat[Local CNN Encoder.\label{fig:local_encoder_details}]{%
\begin{minipage}[t]{0.75\linewidth}

        \centering
        \includegraphics[width=\linewidth]
        {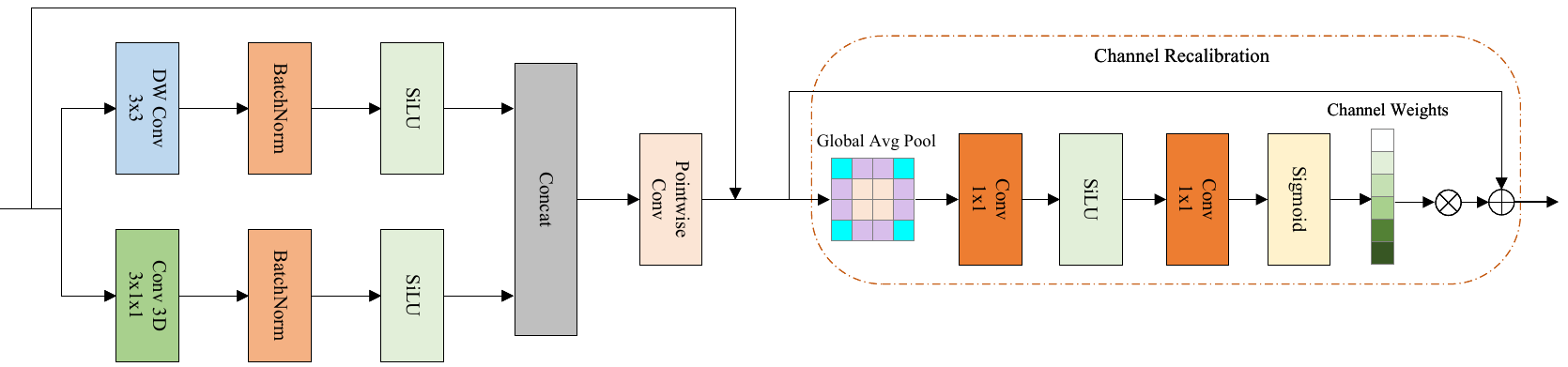}
        \end{minipage}%
}

    \vspace{-0.8mm}
    \caption{Architectures of the asymmetric encoders. (a) Global Mamba Encoder,
    which models broader spatial and spectral dependencies through parallel Mamba
    pathways and gated fusion. (b) Local CNN Encoder, which integrates local
    spectral and spatial convolutional responses with channel recalibration.}
    \label{fig:encoder_details}

\end{figure*}

\subsection{Structured Pruning}
Network pruning aims to reduce model redundancy at the parameter or structural level. Unstructured pruning removes individual connections~\cite{han2016deepcompression}, whereas structured approaches remove
channels or filters to produce compact networks with regular computation~\cite{li2017pruning,he2017channelpruning,luo2017thinet}. Network Slimming
encourages channel sparsity~\cite{liu2017slimming}; Taylor-based criteria
estimate task-related sensitivity~\cite{molchanov2019importance}; and
dependency-aware methods account for structural relationships among pruned
units~\cite{fang2023depgraph}. Group Fisher extends importance evaluation
to groups of coupled channels~\cite{liu2021group}, while prior studies also
analyze the roles of network structure and retraining in compression~\cite{liu2019rethinking}.

For HSI classification, computational efficiency has additionally been
pursued with lightweight convolutional models, morphology-inspired designs,
channel or layer reconfiguration, and efficient attention~\cite{jia2021lightweight,zhou2023adaptive,li2024cloln,zhao2024gscvit}.
These approaches primarily focus on reducing architectural complexity, while conventional structured pruning methods emphasize task sensitivity and structural dependencies. In contrast, KRGF uses the posterior reliability
estimated during global--local fusion as a complementary factor in
Group Fisher scoring. By applying this reliability-weighted criterion to Kalman-coupled channel groups, KRGF connects uncertainty-aware representation learning with structured model compression.

\section{Proposed KGLP-Net}
\label{sec:method}

\subsection{Framework Overview}

For each target pixel at spatial location $p$, two center-aligned
hyperspectral views with different spatial extents are extracted and denoted as $\mathbf{X}_p^g\in\mathbb{R}^{B\times s_g\times s_g}$ and
$\mathbf{X}_p^l\in\mathbb{R}^{B\times s_l\times s_l}$, where $B$ is the
number of spectral bands and $s_g>s_l$. The larger view is processed by a
Global Mamba Encoder $\mathcal{E}_g$ to capture broad contextual information, while the smaller view is
processed by a Local CNN Encoder $\mathcal{E}_l$ to preserve fine-grained spectral--spatial details:
\begin{equation}
\mathbf{G}
=
\mathcal{E}_g(\mathbf{X}_p^g),
\qquad
\mathbf{L}
=
\mathcal{E}_l(\mathbf{X}_p^l),
\label{eq:global_local_repr}
\end{equation}
where both representations have a hidden width of $D$.

As illustrated in Fig.~\ref{fig:overview}, KGLP-Net consists of three main components: asymmetric
global--local representation learning, Kalman-Reliable Fusion (KRF),
and Kalman-Reliable Group Fisher Pruning (KRGF). KRF treats the global representation as a contextual prior and the local representation as an observation, performing a Kalman-inspired feature update based on their learned relative uncertainties. The resulting posterior uncertainty is further exploited by KRGF to derive group-level reliability scores, which are combined with Group Fisher importance to guide the selection of Kalman-coupled channel groups for compact model reconstruction. The fused
representation is finally mapped to class logits through global average pooling and
a linear classifier.

\subsection{Asymmetric Global--Local Representation}
\label{sec:global_local}

The two branches extract complementary spectral--spatial representations through different spatial
scopes and modeling mechanisms. The Global Mamba Encoder operates on the
larger view. After feature embedding and dynamic positional encoding, parallel
spatial and spectral-channel Mamba pathways model contextual dependencies,
and their responses are integrated by gated spatial--spectral fusion. In
contrast, the Local CNN Encoder processes the center-aligned local view using
parallel spectral and spatial convolutions, followed by pointwise feature
integration and channel recalibration. The encoder architectures are illustrated in Fig.~\ref{fig:encoder_details} and described below.

\subsubsection{Global Mamba Encoder}
Following feature embedding and dynamic positional encoding, the global feature is
processed by parallel spatial and spectral Mamba pathways. The spatial pathway
flattens the $H\!\times\!W$ feature grid into a sequence in row-major order and employs unidirectional Mamba scanning to capture spatial contextual dependencies. In parallel, the spectral pathway
applies global average pooling followed by bidirectional Mamba scanning to model complementary spectral-channel dependencies. The resulting features are integrated through spatial--spectral gated fusion, where a learned sigmoid gate adaptively
balances the contributions of the two pathways before residual addition.

\subsubsection{Local CNN Encoder}
The local encoder extracts complementary spectral and spatial features through parallel spectral and spatial convolutional pathways, employing a $3\!\times\!1\!\times\!1$ 3-D convolution and a $3\!\times\!3$ depthwise convolution, respectively. The resulting responses are
concatenated along the channel dimension and integrated through a pointwise
convolution. Channel recalibration then generates channel-wise attention weights through global average pooling, two pointwise
projections with an intermediate SiLU activation, and a sigmoid gate. The
resulting weights adaptively recalibrate the integrated feature before residual
addition, following the general principle of channel-wise feature
recalibration~\cite{hu2018senet}.

In patch-wise HSI classification, the center pixel serves as the prediction target, while its surrounding pixels provide complementary contextual information~\cite{li2023cvssn}.
This distinction motivates assigning different estimation roles to the two views. Specifically, the larger global view captures broader spatial context to provide a context-conditioned estimate of the
target representation, whereas the smaller center-aligned local view preserves fine-grained spectral--spatial information around the target pixel. Accordingly, KRF treats the global representation as a contextual prior and the local representation as an observation, performing a Kalman-inspired update that adaptively adjusts the prior toward the observation based on their learned relative feature-space uncertainties.

\subsection{Kalman-Reliable Fusion}
\label{sec:krf}

KRF performs a Kalman-inspired measurement update in the feature space~\cite{kalman1960new}. Let
$\mathbf{G}\in\mathbb{R}^{D\times H_g\times W_g}$ and
$\mathbf{L}\in\mathbb{R}^{D\times H_l\times W_l}$ denote the global and local
representations, respectively. When their spatial resolutions differ, the global representation $\mathbf{G}$ is resized to match the spatial resolution of $\mathbf{L}$ using bilinear interpolation. For
notational simplicity, the aligned global representation is still denoted by
$\mathbf{G}$ hereafter, such that
$\mathbf{G},\mathbf{L}\in\mathbb{R}^{D\times H\times W}$.

\subsubsection{Interaction-Aware Uncertainty Estimation}
The relative reliability of the global and local representations depends not only on their individual features but also on their interactions. To capture this information, KRF constructs a joint interaction representation:
\begin{equation}
\mathbf{C}
=
[\mathbf{L},\mathbf{G},
 \mathbf{L}\odot\mathbf{G},
 |\mathbf{L}-\mathbf{G}|]
\in
\mathbb{R}^{4D\times H\times W},
\label{eq:krf_interaction}
\end{equation}
where $[\cdot]$ denotes channel-wise concatenation and $\odot$ denotes
element-wise multiplication. Besides the two original representations, the
product term captures their multiplicative interaction and the absolute
difference characterizes their discrepancy.

A lightweight uncertainty estimator $\mathcal{U}$ maps the joint interaction representation $\mathbf{C}$ to two
positive element-wise uncertainty estimates:
\begin{equation}
[\mathbf{P}_g,\mathbf{R}_l]
=
\operatorname{Softplus}
\left(
\operatorname{Split}(\mathcal{U}(\mathbf{C}))
\right)
+\delta,
\label{eq:krf_uncertainty}
\end{equation}
where $\delta=10^{-6}$ ensures positivity and numerical stability.
$\mathbf{P}_g,\mathbf{R}_l
\in\mathbb{R}^{D\times H\times W}$ denote the uncertainty terms associated
with the global prior estimate and local observation, respectively.
In implementation, $\mathcal{U}$ consists of a $1\times1$ projection,
a depthwise $3\times3$ convolution, and a final $1\times1$ projection,
with batch normalization and SiLU activation after the first two operations.
The final $2D$ channels are split into $\mathbf{P}_g$ and $\mathbf{R}_l$.

The uncertainty estimator is trained jointly with the global and local encoders and the classifier using the supervised classification objective. Since the estimated uncertainties determine the Kalman gain in~\eqref{eq:kalman_gain}, which in turn governs the feature update in~\eqref{eq:krf_update}, classification gradients are propagated through the fusion process to optimize $\mathcal{U}$. Consequently, the two positive outputs serve as learned relative feature-space uncertainty estimates that adaptively regulate the global--local posterior update.

\subsubsection{Uncertainty-Guided Posterior Update}
Given the learned uncertainties $\mathbf{P}_g$ and $\mathbf{R}_l$, KRF computes an element-wise
Kalman gain
\begin{equation}
\mathbf{K}
=
\frac{\mathbf{P}_g}
{\mathbf{P}_g+\mathbf{R}_l+\delta},
\label{eq:kalman_gain}
\end{equation}
which is then used to update the global prior toward the local observation:
\begin{equation}
\bar{\mathbf{F}}
=
\mathbf{G}
+
\mathbf{K}\odot(\mathbf{L}-\mathbf{G}).
\label{eq:krf_update}
\end{equation}
Here, a larger $\mathbf{K}$ favors the local observation, whereas a smaller
$\mathbf{K}$ retains more of the global prior. This adaptive update balances the two representations according to their learned relative feature-space uncertainties.

To complement the uncertainty-guided posterior update, KRF incorporates a lightweight residual interaction path that captures additional nonlinear relationships between the global and local representations:
\begin{equation}
\mathbf{F}
=
\Phi_o
\left(
\bar{\mathbf{F}}
+
\operatorname{Dropout}
\left(
\boldsymbol{\alpha}\odot\Phi_i(\mathbf{C})
\right)
\right),
\label{eq:krf_refinement}
\end{equation}
where $\Phi_i$ reduces the $4D$-channel interaction representation to $D$
channels using a $1\times1$ projection followed by a depthwise $3\times3$
convolution, while $\Phi_o$ is a $1\times1$ output projection. Batch normalization and SiLU activation are used within these mappings.
$\boldsymbol{\alpha}\in\mathbb{R}^{D\times1\times1}$ is a learnable
channel-wise scale. $\operatorname{Dropout}(\cdot)$ uses a dropout probability of 0.1 during training.
The resulting $\mathbf{F}$ serves as the fused representation for
classification.

\subsubsection{Posterior Uncertainty}
Based on the learned uncertainties $\mathbf{P}_g$ and $\mathbf{R}_l$, KRF computes the element-wise posterior
uncertainty estimate associated with $\bar{\mathbf{F}}$:
\begin{equation}
\mathbf{P}^{+}
=
\frac{
\mathbf{P}_g\odot\mathbf{R}_l
}{
\mathbf{P}_g+\mathbf{R}_l+\delta
}.
\label{eq:posterior_uncertainty}
\end{equation}
While the final fused representation $\mathbf{F}$ is used for classification, $\mathbf{P}^{+}$ characterizes the estimated feature-space uncertainty associated with the posterior update
$\bar{\mathbf{F}}$. KRGF subsequently aggregates $\mathbf{P}^{+}$ at the channel level and normalizes its reciprocal to derive a relative posterior-reliability score for each Kalman-coupled group.

\subsection{Kalman-Reliable Group Fisher Pruning}
\label{sec:krgf}

Although KRF produces a fused representation and an associated posterior uncertainty estimate, the resulting dense network may still contain redundant feature channels. To address this issue, KRGF incorporates fusion-derived posterior reliability into structured pruning by combining it with Group Fisher task sensitivity.

\subsubsection{Kalman-Coupled Group}
KRF maintains consistent channel indices across the global, local, and fused representations. Accordingly, the $c$-th Kalman-coupled group is defined as
\begin{equation}
\mathcal{S}_c
=
\{\mathbf{G}_c,\mathbf{L}_c,\mathbf{F}_c\},
\qquad
c=1,\ldots,D.
\label{eq:kalman_coupled_group}
\end{equation}
To evaluate group importance, a shared virtual gate $m_c$ is applied to the
$c$-th global and local channels before KRF and to the corresponding fused
channel before classification. Consequently, the sensitivity of $m_c$
reflects the joint task contribution of the complete coupled hidden path
rather than that of an isolated feature channel.

\subsubsection{Group Fisher Importance}
Following the shared-mask formulation of Group Fisher~\cite{liu2021group}, we estimate the task sensitivity of each coupled group
using the squared gradient of the classification objective with respect to its
shared virtual gate. Let $M$ denote the number of calibration mini-batches.
The importance of $\mathcal{S}_c$ is defined as
\begin{equation}
\mathcal{I}_c
=
\frac{1}{M}
\sum_{b=1}^{M}
\left(
\frac{\partial
\mathcal{L}_{\mathrm{cls}}^{(b)}
}{
\partial m_c
}
\right)^2,
\label{eq:group_fisher_importance}
\end{equation}
where $\mathcal{L}_{\mathrm{cls}}^{(b)}$ is the supervised classification
loss on the $b$-th calibration mini-batch. Network parameters are fixed during
this evaluation, and only the virtual gate is used to measure sensitivity.
A larger $\mathcal{I}_c$ indicates a greater influence of the corresponding
coupled path on the classification objective. Since all candidate groups have the same structural width, they are ranked
directly by their Fisher sensitivity.

\subsubsection{Posterior Reliability}
Group Fisher importance measures task sensitivity but does not explicitly account for the reliability of the posterior information. KRGF therefore complements
$\mathcal{I}_c$ with a reliability score derived from the posterior
uncertainty of KRF. For channel $c$, posterior uncertainty is first averaged
over samples and spatial positions within each calibration mini-batch and then
over all $M$ mini-batches:
\begin{equation}
\bar{P}_c^{+}
=
\frac{1}{M}
\sum_{b=1}^{M}
\operatorname{Mean}_{n,h,w}
\left[
P_{b,n,c,h,w}^{+}
\right],
\label{eq:posterior_uncertainty_channel}
\end{equation}
where $n$ indexes samples within mini-batch $b$. Its reciprocal defines the
posterior precision
\begin{equation}
\pi_c
=
\frac{1}{\bar{P}_c^{+}+\delta}.
\label{eq:posterior_precision}
\end{equation}
To obtain a relative reliability measure across coupled groups, the precision
is normalized by its mean over the $D$ hidden groups:
\begin{equation}
\rho_c
=
\frac{\pi_c}
{
\frac{1}{D}\sum_{j=1}^{D}\pi_j+\varepsilon
},
\label{eq:posterior_reliability}
\end{equation}
where $\varepsilon=10^{-12}$ ensures numerical stability. We use
$\rho_c$ as the relative posterior-reliability score of
$\mathcal{S}_c$. A larger $\rho_c$ indicates higher posterior precision
relative to the other coupled groups.

\subsubsection{Reliability-Weighted Group Selection}
Group Fisher importance and posterior reliability capture complementary properties of each Kalman-coupled group. Specifically, Group Fisher importance $\mathcal{I}_c$ measures its sensitivity to the
classification objective, whereas posterior reliability $\rho_c$ reflects the normalized precision of the posterior information produced by the
prior--observation update.
KRGF integrates these two criteria through a multiplicative weighting scheme:
\begin{equation}
Q_c=
\mathcal{I}_c\rho_c.
\label{eq:krgf_score}
\end{equation}
The resulting score jointly accounts for task sensitivity and posterior reliability when ranking coupled groups. Given a target
hidden width $D'$, KRGF retains the Kalman-coupled groups
with the largest scores:
\begin{equation}
\Omega=
\operatorname{TopK}
\left(
\{Q_c\}_{c=1}^{D},D'
\right).
\label{eq:krgf_topk}
\end{equation}
The selected groups define the compact model, ensuring that structured pruning considers both classification relevance and fusion-derived posterior reliability.

\subsection{Training and Compact Reconstruction}
\label{sec:training_compact}

The training and compact reconstruction procedure is summarized in
Algorithm~\ref{alg:kglp_krgf_training}. The dense model (Dense64) is
first optimized using the classification objective. KRGF then performs
progressive compression along
$64\!\rightarrow\!48\!\rightarrow\!40\!\rightarrow\!32$,
producing Compact48, Compact40, and Compact32 at their respective
target widths. At each transition, Group Fisher importance and posterior
reliability are reevaluated on the current network, and the highest-ranked
Kalman-coupled groups determine the reconstructed model. When another
compression step follows, BatchNorm running statistics are reestimated
using one complete pass over the training loader. The model at the
specified target width is finally fine-tuned using the supervised
classification objective.

\begin{algorithm}[H]
\caption{Training and Progressive KRGF Compression}
\label{alg:kglp_krgf_training}
\footnotesize
\begin{algorithmic}[1]
\Require Training loader $\mathcal D_{\mathrm{tr}}$; target width
$D_{\mathrm{tar}}\in\{48,40,32\}$; number of scoring mini-batches $M$
\Ensure Fine-tuned compact model $\mathcal N_{D_{\mathrm{tar}}}$
\Statex \textit{Stage 1: Dense-model training}
\State Train $\mathcal N_{64}$ using $\mathcal L_{\mathrm{cls}}$
\State $\mathcal N\gets\mathcal N_{64}$; $D\gets64$
\Statex \textit{Stage 2: Progressive KRGF compression}
\For{$d\in(48,40,32)$}
    \State Construct groups $\{\mathcal S_c\}_{c=1}^{D}$ and their shared gates $\{m_c\}_{c=1}^{D}$
    \State Freeze model weights and evaluate $M$ scoring mini-batches
    \State Compute $\{\mathcal I_c\}$ using \eqref{eq:group_fisher_importance}
    \State Compute $\{\rho_c\}$ using \eqref{eq:posterior_uncertainty_channel}--\eqref{eq:posterior_reliability}
    \State $Q_c\gets\mathcal I_c\rho_c$ for $c=1,\ldots,D$
    \State $\Omega\gets\operatorname{TopK}(\{Q_c\}_{c=1}^{D},d)$
    \State $\mathcal N\gets\operatorname{Reconstruct}(\mathcal N,\Omega)$; $D\gets d$
    \If{$d=D_{\mathrm{tar}}$}
       \State \textbf{break}
    \EndIf
    \State Reestimate BatchNorm statistics with one pass over $\mathcal D_{\mathrm{tr}}$
\EndFor
\Statex \textit{Stage 3: Compact-model fine-tuning}
\State Fine-tune $\mathcal N$ using $\mathcal L_{\mathrm{cls}}$ on $\mathcal D_{\mathrm{tr}}$
\State $\mathcal N_{D_{\mathrm{tar}}}\gets\mathcal N$
\State \Return $\mathcal N_{D_{\mathrm{tar}}}$
\end{algorithmic}
\end{algorithm}

\section{Experiments}
\label{sec:experiments}

\subsection{Datasets and Evaluation Metrics}
\label{sec:experimental_setup}
\begin{table*}[!t]
\centering
\caption{Numbers of training, validation, and test samples for each class in the four HSI datasets}
\label{tab:dataset_splits}
\begingroup
\fontsize{6.25pt}{7.4pt}\selectfont
\setlength{\tabcolsep}{0.90pt}
\renewcommand{\arraystretch}{1.0}
\definecolor{SplitUP1}{RGB}{255,0,0}
\definecolor{SplitUP2}{RGB}{0,255,0}
\definecolor{SplitUP3}{RGB}{0,0,255}
\definecolor{SplitUP4}{RGB}{255,255,0}
\definecolor{SplitUP5}{RGB}{255,0,255}
\definecolor{SplitUP6}{RGB}{0,255,255}
\definecolor{SplitUP7}{RGB}{200,120,0}
\definecolor{SplitUP8}{RGB}{0,190,100}
\definecolor{SplitUP9}{RGB}{100,0,210}
\definecolor{SplitHOU1}{RGB}{255,0,0}
\definecolor{SplitHOU2}{RGB}{0,255,0}
\definecolor{SplitHOU3}{RGB}{0,0,255}
\definecolor{SplitHOU4}{RGB}{255,255,0}
\definecolor{SplitHOU5}{RGB}{255,0,255}
\definecolor{SplitHOU6}{RGB}{0,255,255}
\definecolor{SplitHOU7}{RGB}{200,120,0}
\definecolor{SplitHOU8}{RGB}{0,190,100}
\definecolor{SplitHOU9}{RGB}{100,0,210}
\definecolor{SplitHOU10}{RGB}{200,0,120}
\definecolor{SplitHOU11}{RGB}{100,210,0}
\definecolor{SplitHOU12}{RGB}{0,115,210}
\definecolor{SplitHOU13}{RGB}{160,80,80}
\definecolor{SplitHOU14}{RGB}{80,160,80}
\definecolor{SplitHOU15}{RGB}{80,80,160}
\definecolor{SplitSAL1}{RGB}{255,0,0}
\definecolor{SplitSAL2}{RGB}{0,255,0}
\definecolor{SplitSAL3}{RGB}{0,0,255}
\definecolor{SplitSAL4}{RGB}{255,255,0}
\definecolor{SplitSAL5}{RGB}{255,0,255}
\definecolor{SplitSAL6}{RGB}{0,255,255}
\definecolor{SplitSAL7}{RGB}{200,120,0}
\definecolor{SplitSAL8}{RGB}{0,190,100}
\definecolor{SplitSAL9}{RGB}{100,0,210}
\definecolor{SplitSAL10}{RGB}{200,0,120}
\definecolor{SplitSAL11}{RGB}{100,210,0}
\definecolor{SplitSAL12}{RGB}{0,115,210}
\definecolor{SplitSAL13}{RGB}{160,80,80}
\definecolor{SplitSAL14}{RGB}{80,160,80}
\definecolor{SplitSAL15}{RGB}{80,80,160}
\definecolor{SplitSAL16}{RGB}{255,90,110}
\definecolor{SplitHAN1}{RGB}{255,0,0}
\definecolor{SplitHAN2}{RGB}{0,255,0}
\definecolor{SplitHAN3}{RGB}{0,0,255}
\definecolor{SplitHAN4}{RGB}{255,255,0}
\definecolor{SplitHAN5}{RGB}{255,0,255}
\definecolor{SplitHAN6}{RGB}{0,255,255}
\definecolor{SplitHAN7}{RGB}{200,120,0}
\definecolor{SplitHAN8}{RGB}{0,190,100}
\definecolor{SplitHAN9}{RGB}{100,0,210}
\definecolor{SplitHAN10}{RGB}{200,0,120}
\definecolor{SplitHAN11}{RGB}{100,210,0}
\definecolor{SplitHAN12}{RGB}{0,115,210}
\definecolor{SplitHAN13}{RGB}{160,80,80}
\definecolor{SplitHAN14}{RGB}{80,160,80}
\definecolor{SplitHAN15}{RGB}{80,80,160}
\definecolor{SplitHAN16}{RGB}{255,90,110}
\newcommand{\KGLPSplitTag}[3]{\cellcolor{#1}\textcolor{#2}{\bfseries #3}}
\begin{tabular*}{\textwidth}{@{\extracolsep{\fill}}
>{\centering\arraybackslash}p{9pt}>{\raggedright\arraybackslash}p{48pt}rrr
@{\hspace{1.5pt}\extracolsep{\fill}}
>{\centering\arraybackslash}p{9pt}>{\raggedright\arraybackslash}p{49pt}rrr
@{\hspace{1.5pt}\extracolsep{\fill}}
>{\centering\arraybackslash}p{9pt}>{\raggedright\arraybackslash}p{53pt}rrr
@{\hspace{1.5pt}\extracolsep{\fill}}
>{\centering\arraybackslash}p{9pt}>{\raggedright\arraybackslash}p{49pt}rrr
@{}}
\toprule
\multicolumn{5}{c}{\textbf{UP (Pavia University)}} &
\multicolumn{5}{c}{\textbf{Houston 2013}} &
\multicolumn{5}{c}{\textbf{Salinas}} &
\multicolumn{5}{c}{\textbf{WHU-Hi-HanChuan}} \\
\cmidrule(lr){1-5}\cmidrule(lr){6-10}\cmidrule(lr){11-15}\cmidrule(lr){16-20}
No. & Class & Train & Val & Test &
No. & Class & Train & Val & Test &
No. & Class & Train & Val & Test &
No. & Class & Train & Val & Test \\
\midrule
\KGLPSplitTag{SplitUP1}{white}{1} & Asphalt & 66 & 331 & 6234 & \KGLPSplitTag{SplitHOU1}{white}{1} & Healthy grass & 62 & 62 & 1127 & \KGLPSplitTag{SplitSAL1}{white}{1} & \resizebox{52pt}{!}{Brocoli green weeds 1} & 20 & 100 & 1889 & \KGLPSplitTag{SplitHAN1}{white}{1} & Strawberry & 447 & 2236 & 42052 \\
\KGLPSplitTag{SplitUP2}{black}{2} & Meadows & 186 & 932 & 17531 & \KGLPSplitTag{SplitHOU2}{black}{2} & Stressed grass & 62 & 62 & 1130 & \KGLPSplitTag{SplitSAL2}{black}{2} & \resizebox{52pt}{!}{Brocoli green weeds 2} & 37 & 186 & 3503 & \KGLPSplitTag{SplitHAN2}{black}{2} & Cowpea & 227 & 1137 & 21389 \\
\KGLPSplitTag{SplitUP3}{white}{3} & Gravel & 20 & 104 & 1975 & \KGLPSplitTag{SplitHOU3}{white}{3} & Synthetic grass & 34 & 34 & 629 & \KGLPSplitTag{SplitSAL3}{white}{3} & Fallow & 19 & 98 & 1859 & \KGLPSplitTag{SplitHAN3}{white}{3} & Soybean & 102 & 514 & 9671 \\
\KGLPSplitTag{SplitUP4}{black}{4} & Trees & 30 & 153 & 2881 & \KGLPSplitTag{SplitHOU4}{black}{4} & Trees & 62 & 62 & 1120 & \KGLPSplitTag{SplitSAL4}{black}{4} & Fallow rough plow & 13 & 69 & 1312 & \KGLPSplitTag{SplitHAN4}{black}{4} & Sorghum & 53 & 267 & 5033 \\
\KGLPSplitTag{SplitUP5}{white}{5} & \resizebox{47pt}{!}{Painted metal sheets} & 13 & 67 & 1265 & \KGLPSplitTag{SplitHOU5}{white}{5} & Soil & 62 & 62 & 1118 & \KGLPSplitTag{SplitSAL5}{white}{5} & Fallow smooth & 26 & 133 & 2519 & \KGLPSplitTag{SplitHAN5}{white}{5} & Water spinach & 12 & 60 & 1128 \\
\KGLPSplitTag{SplitUP6}{black}{6} & Bare Soil & 50 & 251 & 4728 & \KGLPSplitTag{SplitHOU6}{black}{6} & Water & 16 & 16 & 293 & \KGLPSplitTag{SplitSAL6}{black}{6} & Stubble & 39 & 197 & 3723 & \KGLPSplitTag{SplitHAN6}{black}{6} & Watermelon & 45 & 226 & 4262 \\
\KGLPSplitTag{SplitUP7}{white}{7} & Bitumen & 13 & 66 & 1251 & \KGLPSplitTag{SplitHOU7}{white}{7} & Residential & 63 & 63 & 1142 & \KGLPSplitTag{SplitSAL7}{white}{7} & Celery & 35 & 178 & 3366 & \KGLPSplitTag{SplitHAN7}{white}{7} & Greens & 59 & 295 & 5549 \\
\KGLPSplitTag{SplitUP8}{black}{8} & Bricks & 36 & 184 & 3462 & \KGLPSplitTag{SplitHOU8}{black}{8} & Commercial & 62 & 62 & 1120 & \KGLPSplitTag{SplitSAL8}{black}{8} & Grapes untrained & 112 & 563 & 10596 & \KGLPSplitTag{SplitHAN8}{black}{8} & Trees & 179 & 898 & 16901 \\
\KGLPSplitTag{SplitUP9}{white}{9} & Shadows & 9 & 47 & 891 & \KGLPSplitTag{SplitHOU9}{white}{9} & Road & 62 & 62 & 1128 & \KGLPSplitTag{SplitSAL9}{white}{9} & \resizebox{52pt}{!}{Soil vinyard develop} & 62 & 310 & 5831 & \KGLPSplitTag{SplitHAN9}{white}{9} & Grass & 94 & 473 & 8902 \\
 &  &  &  &  & \KGLPSplitTag{SplitHOU10}{white}{10} & Highway & 61 & 61 & 1105 & \KGLPSplitTag{SplitSAL10}{white}{10} & \resizebox{52pt}{!}{Corn senesced-g-w} & 32 & 163 & 3083 & \KGLPSplitTag{SplitHAN10}{white}{10} & Red roof & 105 & 525 & 9886 \\
 &  &  &  &  & \KGLPSplitTag{SplitHOU11}{black}{11} & Railway & 61 & 61 & 1113 & \KGLPSplitTag{SplitSAL11}{black}{11} & \resizebox{52pt}{!}{Lettuce romaine 4wk} & 10 & 53 & 1005 & \KGLPSplitTag{SplitHAN11}{black}{11} & Gray roof & 169 & 845 & 15897 \\
 &  &  &  &  & \KGLPSplitTag{SplitHOU12}{white}{12} & Parking Lot 1 & 61 & 61 & 1111 & \KGLPSplitTag{SplitSAL12}{white}{12} & \resizebox{52pt}{!}{Lettuce romaine 5wk} & 19 & 96 & 1812 & \KGLPSplitTag{SplitHAN12}{white}{12} & Plastic & 36 & 183 & 3460 \\
 &  &  &  &  & \KGLPSplitTag{SplitHOU13}{white}{13} & Parking Lot 2 & 23 & 23 & 423 & \KGLPSplitTag{SplitSAL13}{white}{13} & \resizebox{52pt}{!}{Lettuce romaine 6wk} & 9 & 45 & 862 & \KGLPSplitTag{SplitHAN13}{white}{13} & Bare soil & 91 & 455 & 8570 \\
 &  &  &  &  & \KGLPSplitTag{SplitHOU14}{white}{14} & Tennis Court & 21 & 21 & 386 & \KGLPSplitTag{SplitSAL14}{white}{14} & \resizebox{52pt}{!}{Lettuce romaine 7wk} & 10 & 53 & 1007 & \KGLPSplitTag{SplitHAN14}{white}{14} & Road & 185 & 928 & 17447 \\
 &  &  &  &  & \KGLPSplitTag{SplitHOU15}{white}{15} & Running Track & 33 & 33 & 594 & \KGLPSplitTag{SplitSAL15}{white}{15} & Vinyard untrained & 72 & 363 & 6833 & \KGLPSplitTag{SplitHAN15}{white}{15} & Bright object & 11 & 56 & 1069 \\
 &  &  &  &  &  &  &  &  &  & \KGLPSplitTag{SplitSAL16}{white}{16} & \resizebox{52pt}{!}{Vinyard vertical trellis} & 18 & 90 & 1699 & \KGLPSplitTag{SplitHAN16}{white}{16} & Water & 754 & 3770 & 70877 \\
\midrule
 & \textbf{Total} & \textbf{423} & \textbf{2135} & \textbf{40218} &
 & \textbf{Total} & \textbf{745} & \textbf{745} & \textbf{13539} &
 & \textbf{Total} & \textbf{533} & \textbf{2697} & \textbf{50899} &
 & \textbf{Total} & \textbf{2569} & \textbf{12868} & \textbf{242093} \\
\bottomrule
\end{tabular*}
\endgroup
\end{table*}

\subsubsection{Datasets}
\begin{figure}[!t]
\centering
\includegraphics[width=\columnwidth,keepaspectratio]{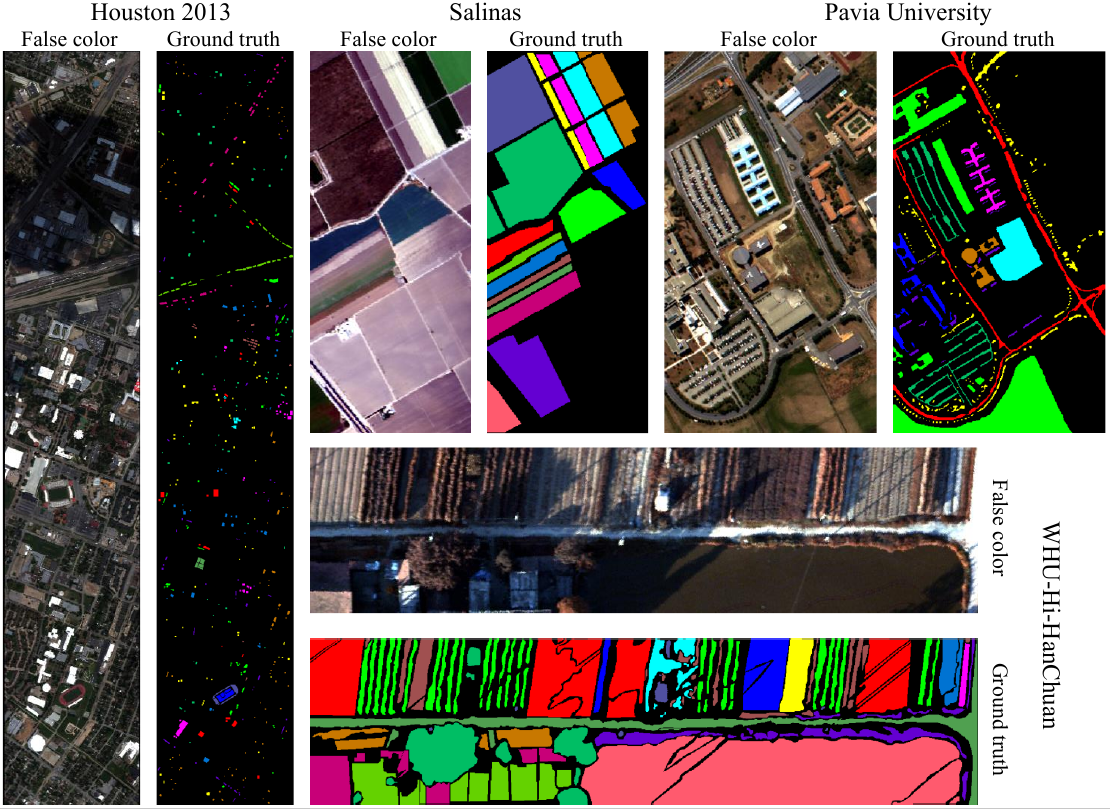}
\caption{False-color composites and corresponding ground-truth maps of
    Houston 2013, Salinas, Pavia University (UP), and WHU-Hi-HanChuan (HanChuan).}
\label{fig:dataset_overview}
\end{figure}

We evaluate KGLP-Net on four public HSI datasets encompassing urban, agricultural, and UAV scenes. Fig.~\ref{fig:dataset_overview} shows their false-color composites and corresponding ground-truth maps.

\indent\textbf{Pavia University (UP):}
The UP image was acquired by the Reflective Optics System Imaging Spectrometer
(ROSIS) over Pavia, Italy. The commonly used image comprises
$610\times340$ pixels and 103 spectral bands, with a spatial resolution of
1.3~m. It contains nine land-cover classes and 42,776 labeled pixels.

\indent\textbf{Houston 2013 (Houston):}
The Houston image was acquired by the ITRES CASI-1500 sensor over the
University of Houston and its surrounding urban area for the 2013 IEEE GRSS
Data Fusion Contest. It comprises $349\times1905$ pixels and 144 spectral
bands, with a spatial resolution of 2.5~m. The ground truth includes
15 classes and 15,029 labeled pixels.

\indent\textbf{Salinas:}
The Salinas image was acquired by the Airborne Visible/Infrared Imaging
Spectrometer (AVIRIS) over agricultural fields in Salinas Valley,
California. It comprises $512\times217$ pixels at a spatial resolution
of approximately 3.7~m. The original 224 bands are reduced to 204 after
removing water-absorption bands. The ground truth includes 16 classes
and 54,129 labeled pixels.

\indent\textbf{WHU-Hi-HanChuan (HanChuan):}
The HanChuan image was acquired by a Headwall Nano-Hyperspec sensor mounted
on a UAV over a rural--urban fringe in Hanchuan, China~\cite{zhong2020whuhi}.
It comprises $1217\times303$ pixels and 274 spectral bands, with a spatial
resolution of approximately 0.109~m. The ground truth includes 16 classes
and 257,530 labeled pixels.

For UP, Salinas, and HanChuan, 1\% of the labeled pixels are used for
training, whereas Houston uses 5\%. An additional 5\% of the labeled
pixels are used for validation on each dataset, and the remainder are
reserved for testing. Background pixels are excluded.
Table~\ref{tab:dataset_splits} presents the class-wise sample counts.

\subsubsection{Evaluation Metrics}
Classification performance is evaluated using overall accuracy (OA),
average accuracy (AA), and Cohen's kappa coefficient
($\kappa$)~\cite{cohen1960coefficient}.
OA measures the proportion of correctly classified test pixels,
AA averages the class-wise accuracies, and $\kappa$ measures agreement
corrected for chance. All experiments are repeated using ten independent
random seeds, with pixel identities resampled while preserving the
class-wise split sizes. Results are reported as mean $\pm$ standard
deviation.

\subsection{Implementation Details}
\label{sec:implementation}
We compare KGLP-Net with ten representative HSI classification methods from three architectural families: CNN-based SSRN~\cite{zhong2018ssrn}, CVSSN~\cite{li2023cvssn}, and
HKNAS~\cite{wang2024hknas}; Transformer-based SSFTT~\cite{sun2022ssftt},
GAHT~\cite{mei2022gaht}, and CMT~\cite{jia2024cmt}; and SSM-based
MambaHSI~\cite{li2024mambahsi}, 3DSS-Mamba~\cite{he2024_3dssmamba},
MambaLG~\cite{pan2025mamba}, and SliMamba~\cite{zhao2026slimamba}. These baselines encompass convolutional spectral--spatial feature extraction, attention-based representation learning, and state-space modeling of contextual dependencies.

For the dense KGLP-Net (Dense64), the hidden width is set to 64 and the Mamba state
dimension is set to 16. The global/local view sizes are set to $21/7$ for UP and Houston,
and $25/9$ for Salinas and HanChuan; their effects are examined in
Sec.~\ref{sec:patchsize}. Dense64 is trained for up to 180 epochs using
AdamW~\cite{loshchilov2019adamw}, with an initial learning rate of
$5\times10^{-4}$, weight decay of 0.05, and batch size of 64. Cosine
annealing~\cite{loshchilov2017sgdr} schedules the learning rate, and
Poly-1 loss~\cite{leng2022polyloss} provides the classification objective.
After the chosen target width is reached through KRGF compression, the reconstructed
compact model is fine-tuned for up to 80 epochs at an initial learning rate of
$2\times10^{-4}$ using the same optimizer and scheduler. All experiments
are implemented in PyTorch, and the dense-to-compact reconstruction procedure is
described in Sec.~\ref{sec:training_compact}.

\subsection{Comparison With Representative Methods}
\begin{table*}[!t]
\centering
\begingroup
\begin{minipage}[t]{\linewidth}
\vspace{0pt}
\centering
\setlength{\abovecaptionskip}{0pt}
\setlength{\belowcaptionskip}{2.5pt}
\captionof{table}{Class-wise and overall classification performance on UP}
\label{tab:classwise_up}
\setlength{\tabcolsep}{1.05pt}
\renewcommand{\arraystretch}{1.05}
\resizebox{\linewidth}{!}{%
\begin{tabular}{l|ccc|ccc|ccccc}
\hline
& \multicolumn{3}{|c|}{CNN-Based} & \multicolumn{3}{c|}{Transformer-Based} & \multicolumn{5}{c}{SSM-Based} \\[-0.15ex]
\cmidrule(lr){2-4}\cmidrule(lr){5-7}\cmidrule(lr){8-12}
Class & \shortstack{CVSSN\\[-0.15ex]\scriptsize(TCSVT~2023)}  & \shortstack{SSRN\\[-0.15ex]\scriptsize(TGRS~2018)} & \shortstack{HKNAS3D\\[-0.15ex]\scriptsize(TNNLS~2024)} & \shortstack{SSFTT\\[-0.15ex]\scriptsize(TGRS~2022)} & \shortstack{GAHT\\[-0.15ex]\scriptsize(TGRS~2022)} & \shortstack{CMT\\[-0.15ex]\scriptsize(TGRS~2024)} & \shortstack{MambaHSI\\[-0.15ex]\scriptsize(TGRS~2024)} & \shortstack{3DSS-Mamba\\[-0.15ex]\scriptsize(TGRS~2024)} & \shortstack{MambaLG\\[-0.15ex]\scriptsize(TGRS~2025)} & \shortstack{SliMamba\\[-0.15ex]\scriptsize(TGRS~2026)} & \shortstack{Compact48\\[-0.15ex]\scriptsize(Proposed)} \\
\hline
1 Asphalt & 97.06$\pm$1.43 & \textbf{99.45$\pm$0.48} & 95.53$\pm$1.11 & 93.51$\pm$2.39 & 96.12$\pm$1.26 & 81.78$\pm$18.24 & 94.12$\pm$2.27 & 88.17$\pm$2.64 & 97.67$\pm$0.94 & 97.23$\pm$1.04 & \underline{99.04$\pm$0.60} \\
2 Meadows & 99.51$\pm$0.45 & 99.67$\pm$0.26 & 97.10$\pm$1.30 & 97.08$\pm$1.70 & 99.50$\pm$0.26 & 85.01$\pm$10.93 & 97.59$\pm$1.04 & 96.90$\pm$0.67 & 99.66$\pm$0.22 & \underline{99.69$\pm$0.13} & \textbf{99.85$\pm$0.17} \\
3 Gravel & \underline{95.80$\pm$2.75} & 94.22$\pm$2.20 & 88.33$\pm$5.22 & 90.51$\pm$1.76 & 88.76$\pm$3.83 & 72.52$\pm$19.27 & 90.46$\pm$4.30 & 75.34$\pm$6.33 & 90.65$\pm$3.84 & 83.97$\pm$7.25 & \textbf{98.09$\pm$0.90} \\
4 Trees & 98.46$\pm$0.54 & 96.48$\pm$1.73 & 95.81$\pm$2.31 & \textbf{98.51$\pm$0.83} & 96.89$\pm$1.32 & 94.24$\pm$2.60 & 92.86$\pm$3.47 & 86.27$\pm$4.32 & 92.00$\pm$3.05 & 95.85$\pm$1.27 & \underline{98.50$\pm$0.51} \\
5 Painted metal sheets & \underline{99.98$\pm$0.05} & \textbf{100.00$\pm$0.00} & 99.75$\pm$0.43 & \textbf{100.00$\pm$0.00} & \textbf{100.00$\pm$0.00} & 98.32$\pm$1.88 & \textbf{100.00$\pm$0.00} & 97.17$\pm$2.83 & 99.85$\pm$0.27 & 99.85$\pm$0.22 & \textbf{100.00$\pm$0.00} \\
6 Bare Soil & 98.72$\pm$1.73 & 99.34$\pm$0.79 & 96.55$\pm$2.18 & 97.87$\pm$1.31 & 98.37$\pm$1.05 & 79.15$\pm$25.84 & 99.00$\pm$0.80 & 80.07$\pm$5.35 & \underline{99.35$\pm$0.81} & 98.21$\pm$1.49 & \textbf{99.97$\pm$0.05} \\
7 Bitumen & 96.61$\pm$2.51 & \underline{99.32$\pm$0.87} & 90.58$\pm$5.64 & 93.33$\pm$3.99 & 95.22$\pm$1.99 & 82.02$\pm$24.63 & 96.62$\pm$2.62 & 62.89$\pm$9.02 & 89.28$\pm$6.98 & 98.06$\pm$1.92 & \textbf{99.91$\pm$0.19} \\
8 Self-Blocking Bricks & 96.32$\pm$2.02 & \underline{96.85$\pm$1.70} & 91.00$\pm$5.10 & 92.85$\pm$3.86 & 93.96$\pm$1.82 & 72.83$\pm$22.34 & 95.79$\pm$1.62 & 89.05$\pm$3.99 & 94.47$\pm$2.80 & 91.61$\pm$3.97 & \textbf{98.70$\pm$0.81} \\
9 Shadows & \underline{99.71$\pm$0.39} & \textbf{99.89$\pm$0.16} & 99.38$\pm$0.50 & 99.44$\pm$0.64 & 99.61$\pm$0.42 & 94.89$\pm$3.84 & 99.37$\pm$0.67 & 77.13$\pm$19.71 & 94.89$\pm$3.78 & 96.52$\pm$1.28 & 99.46$\pm$1.19 \\
\hline
\multicolumn{1}{c|}{OA} & 98.44$\pm$0.23 & \underline{98.86$\pm$0.16} & 95.67$\pm$0.88 & 96.06$\pm$1.07 & 97.54$\pm$0.44 & 83.37$\pm$4.94 & 96.46$\pm$0.46 & 89.59$\pm$1.13 & 97.46$\pm$0.31 & 97.27$\pm$0.39 & \textbf{99.45$\pm$0.12} \\
\multicolumn{1}{c|}{AA} & 98.02$\pm$0.45 & \underline{98.36$\pm$0.28} & 94.89$\pm$0.83 & 95.90$\pm$0.71 & 96.49$\pm$0.70 & 84.53$\pm$3.60 & 96.20$\pm$0.87 & 83.66$\pm$1.96 & 95.31$\pm$0.81 & 95.66$\pm$0.60 & \textbf{99.28$\pm$0.12} \\
\multicolumn{1}{c|}{$\kappa\times100$} & 97.93$\pm$0.31 & \underline{98.49$\pm$0.21} & 94.29$\pm$1.14 & 94.81$\pm$1.38 & 96.74$\pm$0.58 & 78.53$\pm$5.87 & 96.38$\pm$2.05 & 86.16$\pm$1.50 & 96.62$\pm$0.41 & 96.39$\pm$0.51 & \textbf{99.28$\pm$0.16} \\
\hline
\end{tabular}}
\end{minipage}
\endgroup
\end{table*}
\begin{figure*}[!t]
\centering
\appqualportraitfigure{UP}
{Classification maps on UP (corresponding to Table~\ref{tab:classwise_up}).}
{fig:app_qual_up}
\end{figure*}
\begin{table*}[!t]
\centering
\begingroup
\begin{minipage}[t]{\linewidth}
\vspace{0pt}
\centering
\setlength{\abovecaptionskip}{0pt}
\setlength{\belowcaptionskip}{2.5pt}
\captionof{table}{Class-wise and overall classification performance on Houston}
\label{tab:classwise_houston}
\setlength{\tabcolsep}{1.05pt}
\renewcommand{\arraystretch}{1.05}
\resizebox{\linewidth}{!}{%
\begin{tabular}{l|ccc|ccc|ccccc}
\hline
& \multicolumn{3}{|c|}{CNN-Based} & \multicolumn{3}{c|}{Transformer-Based} & \multicolumn{5}{c}{SSM-Based} \\[-0.15ex]
\cmidrule(lr){2-4}\cmidrule(lr){5-7}\cmidrule(lr){8-12}
Class & \shortstack{CVSSN\\[-0.15ex]\scriptsize(TCSVT~2023)}  & \shortstack{SSRN\\[-0.15ex]\scriptsize(TGRS~2018)} & \shortstack{HKNAS3D\\[-0.15ex]\scriptsize(TNNLS~2024)} & \shortstack{SSFTT\\[-0.15ex]\scriptsize(TGRS~2022)} & \shortstack{GAHT\\[-0.15ex]\scriptsize(TGRS~2022)} & \shortstack{CMT\\[-0.15ex]\scriptsize(TGRS~2024)} & \shortstack{MambaHSI\\[-0.15ex]\scriptsize(TGRS~2024)} & \shortstack{3DSS-Mamba\\[-0.15ex]\scriptsize(TGRS~2024)} & \shortstack{MambaLG\\[-0.15ex]\scriptsize(TGRS~2025)} & \shortstack{SliMamba\\[-0.15ex]\scriptsize(TGRS~2026)} & \shortstack{Compact48\\[-0.15ex]\scriptsize(Proposed)} \\
\hline
1 Healthy grass & 98.28$\pm$2.73 & \textbf{99.06$\pm$0.84} & 96.40$\pm$3.41 & 97.79$\pm$2.81 & 98.80$\pm$0.94 & 96.67$\pm$2.37 & 95.37$\pm$4.53 & 94.23$\pm$3.03 & 98.15$\pm$1.91 & 97.35$\pm$1.76 & \underline{98.91$\pm$1.53} \\
2 Stressed grass & 98.63$\pm$1.09 & \underline{99.09$\pm$0.92} & 97.44$\pm$2.10 & 98.22$\pm$1.08 & 98.18$\pm$1.41 & 92.70$\pm$7.47 & 97.27$\pm$1.42 & 94.06$\pm$3.08 & \textbf{99.30$\pm$0.53} & 98.48$\pm$0.55 & 98.02$\pm$1.68 \\
3 Synthetic grass & 99.92$\pm$0.19 & \textbf{100.00$\pm$0.00} & 99.87$\pm$0.25 & \textbf{100.00$\pm$0.00} & 99.94$\pm$0.11 & \underline{99.98$\pm$0.05} & 99.70$\pm$0.44 & 98.36$\pm$1.92 & 99.05$\pm$0.79 & 99.90$\pm$0.16 & \textbf{100.00$\pm$0.00} \\
4 Trees & 94.41$\pm$0.58 & \textbf{99.58$\pm$0.49} & 94.88$\pm$1.81 & 94.96$\pm$1.78 & 96.80$\pm$2.61 & 97.39$\pm$2.64 & 97.12$\pm$2.32 & 92.18$\pm$3.38 & 98.21$\pm$1.23 & \underline{98.74$\pm$1.04} & 97.12$\pm$1.68 \\
5 Soil & 99.90$\pm$0.24 & 99.85$\pm$0.46 & 99.11$\pm$2.04 & 99.76$\pm$0.43 & \underline{99.97$\pm$0.06} & 99.67$\pm$0.82 & 99.92$\pm$0.13 & 99.19$\pm$0.67 & 99.93$\pm$0.10 & 99.94$\pm$0.19 & \textbf{99.98$\pm$0.04} \\
6 Water & 98.36$\pm$4.69 & 95.22$\pm$5.44 & 98.02$\pm$4.50 & 97.54$\pm$4.36 & 90.58$\pm$3.79 & \textbf{98.63$\pm$3.55} & \underline{98.46$\pm$1.48} & 89.90$\pm$5.79 & 95.70$\pm$4.05 & 97.95$\pm$4.85 & 95.73$\pm$5.61 \\
7 Residential & 95.46$\pm$1.88 & \textbf{98.05$\pm$0.53} & 91.89$\pm$3.03 & 90.72$\pm$3.31 & 93.61$\pm$2.47 & 92.06$\pm$3.46 & 92.91$\pm$4.49 & 84.77$\pm$5.38 & 96.42$\pm$2.23 & 96.70$\pm$2.27 & \underline{97.78$\pm$1.21} \\
8 Commercial & 91.62$\pm$2.35 & 92.23$\pm$2.77 & 87.56$\pm$3.05 & 90.91$\pm$2.97 & 91.17$\pm$3.37 & 88.70$\pm$5.60 & 79.86$\pm$5.57 & 85.98$\pm$3.18 & 91.20$\pm$2.66 & \underline{92.29$\pm$2.58} & \textbf{92.78$\pm$2.11} \\
9 Road & \underline{95.00$\pm$2.78} & 93.64$\pm$2.83 & 88.21$\pm$3.60 & 90.73$\pm$2.88 & 91.94$\pm$2.47 & 78.68$\pm$16.31 & 90.64$\pm$2.21 & 88.22$\pm$4.17 & 92.89$\pm$3.77 & 94.71$\pm$2.44 & \textbf{97.37$\pm$1.78} \\
10 Highway & \underline{99.08$\pm$1.52} & 98.85$\pm$1.46 & 95.20$\pm$3.13 & 96.77$\pm$2.44 & 97.56$\pm$1.69 & 81.10$\pm$22.36 & 95.89$\pm$2.57 & 93.38$\pm$4.48 & 98.24$\pm$1.87 & 98.88$\pm$1.41 & \textbf{99.81$\pm$0.38} \\
11 Railway & 97.90$\pm$1.62 & 97.61$\pm$1.30 & 88.48$\pm$4.37 & 87.39$\pm$4.54 & 96.01$\pm$1.32 & 68.24$\pm$16.54 & 90.92$\pm$4.95 & 93.02$\pm$2.50 & 96.86$\pm$2.50 & \textbf{99.26$\pm$0.89} & \underline{98.10$\pm$0.98} \\
12 Parking Lot 1 & 96.55$\pm$1.07 & \underline{97.20$\pm$2.76} & 92.42$\pm$3.41 & 94.23$\pm$1.91 & 95.53$\pm$1.59 & 69.69$\pm$15.19 & 90.35$\pm$3.47 & 91.51$\pm$4.66 & 96.80$\pm$2.67 & 95.18$\pm$3.28 & \textbf{98.12$\pm$1.62} \\
13 Parking Lot 2 & 97.87$\pm$2.14 & 94.89$\pm$1.79 & 95.32$\pm$1.22 & 96.12$\pm$1.28 & 96.05$\pm$1.26 & 95.13$\pm$2.50 & \underline{98.11$\pm$2.15} & 89.50$\pm$3.75 & 94.00$\pm$2.12 & 92.10$\pm$2.92 & \textbf{98.44$\pm$2.00} \\
14 Tennis Court & \textbf{100.00$\pm$0.00} & \textbf{100.00$\pm$0.00} & 99.82$\pm$0.33 & \textbf{100.00$\pm$0.00} & 99.82$\pm$0.28 & \underline{99.84$\pm$0.39} & \textbf{100.00$\pm$0.00} & 98.99$\pm$0.76 & 99.22$\pm$1.11 & \textbf{100.00$\pm$0.00} & \textbf{100.00$\pm$0.00} \\
15 Running Track & \textbf{100.00$\pm$0.00} & \textbf{100.00$\pm$0.00} & 99.90$\pm$0.22 & 99.85$\pm$0.45 & \textbf{100.00$\pm$0.00} & 98.13$\pm$3.73 & 97.61$\pm$1.42 & 98.55$\pm$1.31 & 99.81$\pm$0.21 & \underline{99.98$\pm$0.05} & \textbf{100.00$\pm$0.00} \\
\hline
\multicolumn{1}{c|}{OA} & 97.14$\pm$0.40 & \underline{97.68$\pm$0.41} & 94.13$\pm$0.77 & 94.97$\pm$0.81 & 96.31$\pm$0.60 & 88.58$\pm$2.10 & 93.97$\pm$1.08 & 92.36$\pm$1.52 & 96.99$\pm$0.53 & 97.34$\pm$0.42 & \textbf{98.03$\pm$0.21} \\
\multicolumn{1}{c|}{AA} & 97.53$\pm$0.41 & \underline{97.69$\pm$0.53} & 94.97$\pm$0.66 & 95.67$\pm$0.75 & 96.40$\pm$0.61 & 90.44$\pm$1.65 & 94.94$\pm$0.93 & 92.79$\pm$1.48 & 97.05$\pm$0.42 & 97.43$\pm$0.52 & \textbf{98.14$\pm$0.42} \\
\multicolumn{1}{c|}{$\kappa\times100$} & 96.91$\pm$0.43 & \underline{97.49$\pm$0.44} & 93.66$\pm$0.83 & 94.56$\pm$0.87 & 96.01$\pm$0.65 & 87.66$\pm$2.27 & 94.00$\pm$2.31 & 91.74$\pm$1.65 & 96.75$\pm$0.57 & 97.13$\pm$0.46 & \textbf{97.87$\pm$0.23} \\
\hline
\end{tabular}}
\end{minipage}
\endgroup
\end{table*}
\begin{figure*}[!t]
\centering
\appqualhoustonfigure{Houston}
{Classification maps on Houston 2013 (corresponding to Table~\ref{tab:classwise_houston}).}
{fig:app_qual_houston}
\end{figure*}
\begin{table*}[!t]
\centering
\begingroup
\begin{minipage}[t]{\linewidth}
\vspace{0pt}
\centering
\setlength{\abovecaptionskip}{0pt}
\setlength{\belowcaptionskip}{2.5pt}
\captionof{table}{Class-wise and overall classification performance on Salinas}
\label{tab:classwise_salinas}
\setlength{\tabcolsep}{1.05pt}
\renewcommand{\arraystretch}{1.05}
\resizebox{\linewidth}{!}{%
\begin{tabular}{l|ccc|ccc|ccccc}
\hline
& \multicolumn{3}{|c|}{CNN-Based} & \multicolumn{3}{c|}{Transformer-Based} & \multicolumn{5}{c}{SSM-Based} \\[-0.15ex]
\cmidrule(lr){2-4}\cmidrule(lr){5-7}\cmidrule(lr){8-12}
Class & \shortstack{CVSSN\\[-0.15ex]\scriptsize(TCSVT~2023)}  & \shortstack{SSRN\\[-0.15ex]\scriptsize(TGRS~2018)} & \shortstack{HKNAS3D\\[-0.15ex]\scriptsize(TNNLS~2024)} & \shortstack{SSFTT\\[-0.15ex]\scriptsize(TGRS~2022)} & \shortstack{GAHT\\[-0.15ex]\scriptsize(TGRS~2022)} & \shortstack{CMT\\[-0.15ex]\scriptsize(TGRS~2024)} & \shortstack{MambaHSI\\[-0.15ex]\scriptsize(TGRS~2024)} & \shortstack{3DSS-Mamba\\[-0.15ex]\scriptsize(TGRS~2024)} & \shortstack{MambaLG\\[-0.15ex]\scriptsize(TGRS~2025)} & \shortstack{SliMamba\\[-0.15ex]\scriptsize(TGRS~2026)} & \shortstack{Compact48\\[-0.15ex]\scriptsize(Proposed)} \\
\hline
1 Brocoli green weeds 1 & 99.95$\pm$0.11 & \textbf{100.00$\pm$0.00} & 98.96$\pm$1.61 & 99.47$\pm$0.74 & 99.28$\pm$1.26 & 99.80$\pm$0.29 & \underline{99.97$\pm$0.08} & 98.58$\pm$1.56 & 99.63$\pm$0.48 & \underline{99.97$\pm$0.05} & \underline{99.97$\pm$0.10} \\
2 Brocoli green weeds 2 & 99.96$\pm$0.04 & \textbf{100.00$\pm$0.00} & 99.39$\pm$1.35 & 99.98$\pm$0.05 & 99.85$\pm$0.26 & 99.65$\pm$0.52 & 98.43$\pm$3.45 & 97.04$\pm$1.96 & \underline{99.99$\pm$0.03} & 99.96$\pm$0.08 & \textbf{100.00$\pm$0.00} \\
3 Fallow & 99.70$\pm$0.28 & 98.49$\pm$2.77 & 97.70$\pm$1.69 & 97.73$\pm$2.06 & 97.64$\pm$2.43 & 96.63$\pm$9.10 & 75.34$\pm$20.96 & 92.79$\pm$5.21 & 93.51$\pm$9.46 & \underline{99.87$\pm$0.20} & \textbf{99.96$\pm$0.13} \\
4 Fallow rough plow & 99.35$\pm$0.48 & \textbf{99.92$\pm$0.20} & 98.11$\pm$2.04 & 99.73$\pm$0.65 & 99.60$\pm$0.50 & 94.67$\pm$8.32 & \underline{99.90$\pm$0.10} & 94.56$\pm$3.45 & 99.27$\pm$0.40 & 99.27$\pm$0.74 & \textbf{99.92$\pm$0.12} \\
5 Fallow smooth & 95.54$\pm$3.07 & 97.58$\pm$1.31 & 97.46$\pm$1.75 & 97.25$\pm$1.95 & 97.36$\pm$1.43 & 96.67$\pm$2.50 & 95.44$\pm$3.31 & 96.17$\pm$2.08 & 97.36$\pm$1.63 & \underline{97.84$\pm$1.73} & \textbf{98.22$\pm$0.89} \\
6 Stubble & 99.95$\pm$0.09 & \underline{99.99$\pm$0.02} & 99.79$\pm$0.33 & 99.54$\pm$0.66 & 99.95$\pm$0.12 & 99.88$\pm$0.21 & 99.78$\pm$0.35 & 99.40$\pm$0.53 & \textbf{100.00$\pm$0.01} & 99.96$\pm$0.07 & 99.97$\pm$0.09 \\
7 Celery & 99.83$\pm$0.20 & \textbf{100.00$\pm$0.01} & 98.53$\pm$1.56 & 99.84$\pm$0.16 & 99.83$\pm$0.20 & 99.77$\pm$0.30 & 99.57$\pm$0.59 & 97.77$\pm$0.93 & 99.82$\pm$0.20 & 99.66$\pm$0.32 & \underline{99.98$\pm$0.03} \\
8 Grapes untrained & 87.97$\pm$1.62 & 93.88$\pm$3.10 & 81.61$\pm$3.28 & 79.79$\pm$4.43 & 90.65$\pm$1.60 & 83.88$\pm$9.47 & 85.59$\pm$6.85 & 92.42$\pm$2.47 & \textbf{97.49$\pm$1.44} & 94.40$\pm$1.58 & \underline{95.15$\pm$2.63} \\
9 Soil vinyard develop & 99.47$\pm$0.43 & \underline{99.98$\pm$0.06} & 98.21$\pm$2.29 & 99.71$\pm$0.31 & 99.92$\pm$0.14 & 99.49$\pm$1.38 & 98.37$\pm$4.89 & 98.79$\pm$0.74 & 99.95$\pm$0.08 & \underline{99.98$\pm$0.02} & \textbf{100.00$\pm$0.00} \\
10 Corn senesced green weeds & \underline{98.58$\pm$0.80} & 96.57$\pm$1.19 & 97.30$\pm$1.66 & 97.33$\pm$1.49 & 95.76$\pm$2.80 & 95.35$\pm$4.12 & 83.17$\pm$6.75 & 92.86$\pm$2.50 & 96.09$\pm$1.71 & \textbf{98.69$\pm$0.69} & 97.99$\pm$1.16 \\
11 Lettuce romaine 4wk & 99.31$\pm$1.15 & 98.44$\pm$1.37 & 96.65$\pm$2.86 & 94.89$\pm$4.17 & 96.93$\pm$2.72 & 98.61$\pm$2.38 & 95.32$\pm$10.08 & 87.14$\pm$3.13 & 91.99$\pm$4.37 & \underline{99.44$\pm$0.72} & \textbf{99.84$\pm$0.35} \\
12 Lettuce romaine 5wk & 99.65$\pm$0.32 & \textbf{100.00$\pm$0.00} & 99.06$\pm$1.51 & 99.64$\pm$0.36 & 99.88$\pm$0.22 & 94.92$\pm$5.42 & 91.35$\pm$11.73 & 92.45$\pm$3.83 & 99.07$\pm$0.94 & 99.61$\pm$0.44 & \underline{99.89$\pm$0.10} \\
13 Lettuce romaine 6wk & 99.30$\pm$1.02 & 99.06$\pm$1.42 & 99.18$\pm$2.07 & 96.90$\pm$4.16 & 98.13$\pm$2.39 & 92.38$\pm$12.78 & 96.64$\pm$5.98 & 96.16$\pm$4.81 & 93.70$\pm$11.52 & \underline{99.57$\pm$0.66} & \textbf{99.74$\pm$0.32} \\
14 Lettuce romaine 7wk & \underline{99.40$\pm$0.61} & 98.28$\pm$1.30 & 98.88$\pm$1.59 & 97.02$\pm$3.28 & 98.03$\pm$1.61 & 98.14$\pm$2.86 & 95.28$\pm$5.85 & 86.78$\pm$8.40 & 93.76$\pm$4.49 & 98.18$\pm$1.80 & \textbf{99.63$\pm$0.24} \\
15 Vinyard untrained & 92.49$\pm$2.02 & 92.82$\pm$2.53 & 87.13$\pm$6.47 & 86.12$\pm$3.80 & 89.40$\pm$2.25 & 82.63$\pm$13.91 & 93.61$\pm$7.83 & 91.58$\pm$3.20 & \underline{95.34$\pm$3.34} & 95.18$\pm$2.04 & \textbf{98.09$\pm$2.16} \\
16 Vinyard vertical trellis & \textbf{99.12$\pm$0.64} & 97.65$\pm$3.58 & \underline{98.57$\pm$1.35} & 97.91$\pm$1.23 & 95.92$\pm$5.04 & 97.89$\pm$1.46 & 98.47$\pm$2.71 & 90.34$\pm$5.28 & 95.07$\pm$5.91 & 98.05$\pm$1.87 & 98.56$\pm$4.00 \\
\hline
\multicolumn{1}{c|}{OA} & 95.99$\pm$0.37 & 97.21$\pm$0.45 & 93.44$\pm$0.57 & 93.15$\pm$1.12 & 95.81$\pm$0.31 & 93.05$\pm$1.48 & 93.04$\pm$2.80 & 94.51$\pm$1.03 & 97.61$\pm$0.60 & \underline{97.81$\pm$0.55} & \textbf{98.45$\pm$0.83} \\
\multicolumn{1}{c|}{AA} & 98.10$\pm$0.29 & 98.29$\pm$0.32 & 96.66$\pm$0.34 & 96.43$\pm$0.63 & 97.38$\pm$0.44 & 95.65$\pm$1.06 & 94.14$\pm$2.87 & 94.05$\pm$1.22 & 97.00$\pm$0.74 & \underline{98.73$\pm$0.31} & \textbf{99.18$\pm$0.47} \\
\multicolumn{1}{c|}{$\kappa\times100$} & 95.55$\pm$0.42 & 96.90$\pm$0.49 & 92.71$\pm$0.63 & 92.39$\pm$1.25 & 95.34$\pm$0.34 & 92.27$\pm$1.64 & 92.67$\pm$3.53 & 93.88$\pm$1.15 & 97.34$\pm$0.66 & \underline{97.56$\pm$0.62} & \textbf{98.27$\pm$0.92} \\
\hline
\end{tabular}}
\end{minipage}
\endgroup
\end{table*}
\begin{figure*}[!t]
\centering
\appqualportraitfigure{Salinas}
{Classification maps on Salinas (corresponding to Table~\ref{tab:classwise_salinas}).}
{fig:app_qual_salinas}
\end{figure*}
\begin{table*}[!t]
\centering
\begingroup
\begin{minipage}[t]{\linewidth}
\vspace{0pt}
\centering
\setlength{\abovecaptionskip}{0pt}
\setlength{\belowcaptionskip}{2.5pt}
\captionof{table}{Class-wise and overall classification performance on HanChuan}
\label{tab:classwise_hanchuan}
\setlength{\tabcolsep}{1.05pt}
\renewcommand{\arraystretch}{1.05}
\resizebox{\linewidth}{!}{%
\begin{tabular}{l|ccc|ccc|ccccc}
\hline
& \multicolumn{3}{|c|}{CNN-Based} & \multicolumn{3}{c|}{Transformer-Based} & \multicolumn{5}{c}{SSM-Based} \\[-0.15ex]
\cmidrule(lr){2-4}\cmidrule(lr){5-7}\cmidrule(lr){8-12}
Class & \shortstack{CVSSN\\[-0.15ex]\scriptsize(TCSVT~2023)}  & \shortstack{SSRN\\[-0.15ex]\scriptsize(TGRS~2018)} & \shortstack{HKNAS3D\\[-0.15ex]\scriptsize(TNNLS~2024)} & \shortstack{SSFTT\\[-0.15ex]\scriptsize(TGRS~2022)} & \shortstack{GAHT\\[-0.15ex]\scriptsize(TGRS~2022)} & \shortstack{CMT\\[-0.15ex]\scriptsize(TGRS~2024)} & \shortstack{MambaHSI\\[-0.15ex]\scriptsize(TGRS~2024)} & \shortstack{3DSS-Mamba\\[-0.15ex]\scriptsize(TGRS~2024)} & \shortstack{MambaLG\\[-0.15ex]\scriptsize(TGRS~2025)} & \shortstack{SliMamba\\[-0.15ex]\scriptsize(TGRS~2026)} & \shortstack{Compact48\\[-0.15ex]\scriptsize(Proposed)} \\
\hline
1 Strawberry & 98.05$\pm$0.53 & 97.92$\pm$0.52 & 89.94$\pm$2.89 & 95.20$\pm$1.14 & 97.81$\pm$1.07 & 89.58$\pm$3.75 & 89.57$\pm$5.74 & 93.88$\pm$1.05 & \underline{99.00$\pm$0.34} & 96.75$\pm$0.60 & \textbf{99.27$\pm$0.27} \\
2 Cowpea & 97.41$\pm$0.59 & 94.33$\pm$0.94 & 88.13$\pm$5.11 & 90.88$\pm$1.37 & 96.62$\pm$0.72 & 84.49$\pm$7.95 & 81.69$\pm$4.83 & 86.63$\pm$3.02 & \underline{98.07$\pm$0.64} & 92.14$\pm$1.39 & \textbf{99.03$\pm$0.65} \\
3 Soybean & 98.64$\pm$0.68 & 90.40$\pm$3.64 & 93.24$\pm$3.22 & 87.62$\pm$2.17 & 95.73$\pm$1.77 & 86.70$\pm$5.37 & 94.47$\pm$2.88 & 80.01$\pm$5.72 & \underline{99.30$\pm$0.38} & 95.41$\pm$1.04 & \textbf{99.91$\pm$0.07} \\
4 Sorghum & 98.23$\pm$1.16 & 97.65$\pm$0.68 & 96.72$\pm$2.74 & 96.60$\pm$1.21 & 98.07$\pm$1.24 & 96.68$\pm$1.80 & \underline{98.40$\pm$1.64} & 93.44$\pm$1.42 & 96.26$\pm$1.53 & 97.34$\pm$1.13 & \textbf{99.55$\pm$0.43} \\
5 Water spinach & 94.82$\pm$4.10 & 81.17$\pm$7.85 & 93.30$\pm$5.95 & 76.70$\pm$8.00 & 91.14$\pm$5.53 & 78.68$\pm$15.51 & \textbf{99.67$\pm$0.44} & 40.93$\pm$11.70 & 59.39$\pm$33.46 & 93.69$\pm$3.75 & \underline{99.33$\pm$1.30} \\
6 Watermelon & 81.38$\pm$4.98 & 64.45$\pm$5.70 & \underline{81.69$\pm$5.12} & 67.61$\pm$2.37 & 69.90$\pm$4.41 & 62.45$\pm$10.38 & 60.40$\pm$11.74 & 41.32$\pm$5.11 & 78.66$\pm$15.45 & 74.28$\pm$4.65 & \textbf{94.40$\pm$3.01} \\
7 Greens & \underline{95.13$\pm$2.02} & 91.86$\pm$1.59 & 93.28$\pm$3.26 & 88.54$\pm$2.79 & 89.65$\pm$2.87 & 77.13$\pm$16.13 & 94.53$\pm$3.72 & 82.22$\pm$5.40 & 93.67$\pm$2.98 & 90.83$\pm$2.93 & \textbf{99.16$\pm$0.92} \\
8 Trees & 90.13$\pm$3.13 & 92.59$\pm$2.28 & 78.67$\pm$10.66 & 86.97$\pm$2.39 & 90.44$\pm$1.58 & 76.74$\pm$9.17 & 76.97$\pm$6.65 & 79.52$\pm$4.21 & \underline{96.91$\pm$1.24} & 89.83$\pm$0.98 & \textbf{97.97$\pm$0.50} \\
9 Grass & \underline{96.10$\pm$1.29} & 94.49$\pm$1.53 & 89.22$\pm$2.61 & 89.33$\pm$2.79 & 91.24$\pm$2.65 & 87.37$\pm$4.15 & 86.67$\pm$4.87 & 71.19$\pm$3.76 & 95.23$\pm$2.04 & 89.13$\pm$1.95 & \textbf{98.97$\pm$0.58} \\
10 Red roof & \underline{98.46$\pm$0.71} & 97.06$\pm$1.53 & 97.19$\pm$1.19 & 96.03$\pm$1.46 & 98.42$\pm$0.89 & 93.26$\pm$7.36 & 91.76$\pm$3.17 & 92.14$\pm$1.69 & 98.25$\pm$1.16 & 97.38$\pm$1.25 & \textbf{99.38$\pm$0.27} \\
11 Gray roof & 96.28$\pm$2.97 & 96.80$\pm$0.95 & 90.12$\pm$5.01 & 95.72$\pm$0.98 & 95.17$\pm$1.80 & 84.96$\pm$12.75 & 89.89$\pm$6.81 & 94.45$\pm$1.38 & \underline{97.68$\pm$1.34} & 97.15$\pm$0.73 & \textbf{99.18$\pm$0.38} \\
12 Plastic & \underline{91.27$\pm$4.78} & 76.15$\pm$5.45 & 90.59$\pm$5.84 & 78.64$\pm$4.13 & 83.99$\pm$4.13 & 70.53$\pm$19.03 & 84.96$\pm$8.19 & 74.03$\pm$5.93 & 89.89$\pm$11.51 & 89.13$\pm$3.76 & \textbf{98.64$\pm$1.17} \\
13 Bare soil & \underline{90.97$\pm$1.84} & 76.09$\pm$4.48 & 86.61$\pm$3.21 & 77.70$\pm$2.81 & 81.67$\pm$3.66 & 70.23$\pm$9.38 & 78.10$\pm$4.70 & 68.31$\pm$2.35 & 88.86$\pm$2.26 & 78.33$\pm$3.69 & \textbf{94.96$\pm$1.29} \\
14 Road & 94.88$\pm$1.50 & 92.52$\pm$1.76 & 87.04$\pm$4.69 & 89.62$\pm$2.02 & 93.21$\pm$3.06 & 86.94$\pm$8.94 & 91.39$\pm$2.51 & 87.85$\pm$2.84 & \underline{97.55$\pm$0.57} & 91.74$\pm$1.73 & \textbf{98.83$\pm$0.66} \\
15 Bright object & 84.42$\pm$6.46 & 69.80$\pm$9.25 & 85.73$\pm$6.25 & 83.56$\pm$6.64 & 79.46$\pm$9.03 & 81.19$\pm$7.08 & \textbf{94.80$\pm$2.09} & 55.49$\pm$3.64 & 26.70$\pm$27.41 & 79.74$\pm$6.24 & \underline{90.68$\pm$5.19} \\
16 Water & 99.39$\pm$0.20 & 99.54$\pm$0.30 & 97.23$\pm$3.25 & 98.21$\pm$0.31 & 99.45$\pm$0.37 & 99.00$\pm$0.53 & 96.87$\pm$3.23 & 99.27$\pm$0.28 & \textbf{99.82$\pm$0.06} & 99.10$\pm$0.67 & \underline{99.55$\pm$0.23} \\
\hline
\multicolumn{1}{c|}{OA} & 96.68$\pm$0.53 & 94.76$\pm$0.45 & 91.29$\pm$0.79 & 92.67$\pm$0.41 & 95.31$\pm$0.40 & 88.79$\pm$1.83 & 89.85$\pm$1.08 & 89.15$\pm$0.68 & \underline{97.13$\pm$0.77} & 94.54$\pm$0.38 & \textbf{98.94$\pm$0.14} \\
\multicolumn{1}{c|}{AA} & \underline{94.10$\pm$1.14} & 88.30$\pm$1.03 & 89.92$\pm$0.66 & 87.43$\pm$0.96 & 90.75$\pm$0.83 & 82.87$\pm$2.61 & 88.13$\pm$1.16 & 77.54$\pm$1.30 & 88.45$\pm$3.98 & 90.75$\pm$0.65 & \textbf{98.05$\pm$0.42} \\
\multicolumn{1}{c|}{$\kappa\times100$} & 96.11$\pm$0.62 & 93.87$\pm$0.53 & 89.87$\pm$0.90 & 91.43$\pm$0.47 & 94.51$\pm$0.46 & 86.92$\pm$2.12 & 86.67$\pm$3.17 & 87.30$\pm$0.79 & \underline{96.64$\pm$0.90} & 93.61$\pm$0.44 & \textbf{98.76$\pm$0.16} \\
\hline
\end{tabular}}
\end{minipage}
\endgroup
\end{table*}
\begin{figure*}[!t]
\centering
\appqualhanonefigure{HanChuan}
{Classification maps on WHU-Hi-HanChuan (corresponding to Table~\ref{tab:classwise_hanchuan}).}
{fig:app_qual_hanchuan}
\end{figure*}

\begin{table*}[!t]
\centering
\caption{Computational costs of representative HSI classifiers across four
    datasets. Full training time, test inference time, parameter count, and FLOPs are reported
    separately for each dataset}
\label{tab:cross_method_complexity}

\begingroup
\fontsize{6pt}{7.2pt}\selectfont
\setlength{\tabcolsep}{1.0pt}
\renewcommand{\arraystretch}{1.02}

\providecommand{\costmethod}[2]{%
    \shortstack[c]{%
        \mbox{#1}\\
        {\fontsize{5pt}{6pt}\selectfont (#2)}%
    }%
}

\begin{tabular*}{\textwidth}{
@{\extracolsep{\fill}}
cc
!{\hspace{2pt}\vrule width .35pt\hspace{2pt}}
ccc
!{\hspace{2pt}\vrule width .35pt\hspace{2pt}}
ccc
!{\hspace{2pt}\vrule width .35pt\hspace{2pt}}
cccc
!{\hspace{2pt}\vrule width .35pt\hspace{2pt}}
c@{}
}

\toprule
\multirow{2}{*}{Dataset}
& \multirow{2}{*}{Metric}
& \multicolumn{3}{c!{\hspace{2pt}\vrule width .35pt\hspace{2pt}}}{CNN-Based}
& \multicolumn{3}{c!{\hspace{2pt}\vrule width .35pt\hspace{2pt}}}{Transformer-Based}
& \multicolumn{4}{c!{\hspace{2pt}\vrule width .35pt\hspace{2pt}}}{SSM-Based}
& \textbf{Proposed} \\

\cmidrule(lr){3-5}
\cmidrule(lr){6-8}
\cmidrule(lr){9-12}
\cmidrule(l){13-13}

& &
\costmethod{CVSSN}{TCSVT 2023} &
\costmethod{SSRN}{TGRS 2018} &
\costmethod{HKNAS3D}{TNNLS 2024} &
\costmethod{SSFTT}{TGRS 2022} &
\costmethod{GAHT}{TGRS 2022} &
\costmethod{CMT}{TGRS 2024} &
\costmethod{MambaHSI}{TGRS 2024} &
\costmethod{3DSS-Mamba}{TGRS 2024} &
\costmethod{MambaLG}{TGRS 2025} &
\costmethod{SliMamba}{TGRS 2026} &
\costmethod{\textbf{Compact48}}{\textbf{Proposed}} \\

\midrule

\multirow{4}{*}{UP}
& Train Time (s)
& 16.26 & 167.82 & 27.03
& 15.57 & 127.70 & 16.80
& 395.06 & 113.19 & 1984.38 & 40.54
& \textbf{42.57} \\

& Test Time (s)
& 0.23 & 3.09 & 0.53
& 0.18 & 3.70 & 0.65
& 0.58 & 10.33 & 0.27 & 0.86
& \textbf{0.90} \\

& Params (K)
& 247.26 & 216.54 & 98.87
& 484.42 & 927.11 & 249.21
& 428.34 & 10.31 & 186.60 & 13.32
& \textbf{75.31} \\

& FLOPs (M)
& 20.03 & 100.15 & 1.61
& 12.96 & 90.82 & 5.20
& 478.81 & 13.92 & 23.18 & 0.90
& \textbf{13.66} \\

\midrule

\multirow{4}{*}{Houston}
& Train Time (s)
& 19.59 & 252.17 & 57.08
& 20.48 & 63.64 & 19.56
& 4120.51 & 127.24 & 3170.45 & 62.31
& \textbf{32.19} \\

& Test Time (s)
& 0.08 & 0.85 & 0.40
& 0.10 & 0.80 & 0.22
& 1.77 & 3.59 & 0.26 & 0.25
& \textbf{0.46} \\

& Params (K)
& 253.36 & 278.13 & 103.04
& 673.73 & 963.34 & 250.76
& 439.65 & 10.51 & 189.61 & 13.84
& \textbf{79.54} \\

& FLOPs (M)
& 20.47 & 139.05 & 1.82
& 18.18 & 94.31 & 5.20
& 504.24 & 13.92 & 23.67 & 0.93
& \textbf{15.59} \\

\midrule

\multirow{4}{*}{Salinas}
& Train Time (s)
& 25.50 & 382.84 & 37.66
& 25.28 & 84.98 & 19.15
& 140.64 & 154.12 & 375.85 & 59.67
& \textbf{90.53} \\

& Test Time (s)
& 0.38 & 4.96 & 0.69
& 0.35 & 2.78 & 0.76
& 0.26 & 13.20 & 0.25 & 1.23
& \textbf{2.79} \\

& Params (K)
& 264.02 & 401.03 & 108.42
& 1042.44 & 974.93 & 103.36
& 460.34 & 10.54 & 194.80 & 14.51
& \textbf{87.27} \\

& FLOPs (M)
& 21.33 & 216.86 & 2.24
& 28.37 & 95.44 & 5.20
& 553.45 & 13.92 & 24.51 & 0.98
& \textbf{28.35} \\

\midrule

\multirow{4}{*}{HanChuan}
& Train Time (s)
& 96.71 & 853.53 & 136.55
& 96.99 & 256.78 & 63.30
& 459.07 & 397.93 & 476.33 & 268.96
& \textbf{466.95} \\

& Test Time (s)
& 1.98 & 14.60 & 3.33
& 1.90 & 9.38 & 3.19
& 0.59 & 31.33 & 0.19 & 5.32
& \textbf{11.99} \\

& Params (K)
& 270.65 & 477.83 & 111.62
& 1272.84 & 1519.18 & 251.02
& 473.19 & 10.54 & 198.00 & 14.96
& \textbf{92.07} \\

& FLOPs (M)
& 21.88 & 265.48 & 2.50
& 34.74 & 148.75 & 5.20
& 584.18 & 13.92 & 25.02 & 1.02
& \textbf{31.74} \\

\bottomrule
\end{tabular*}

\endgroup
\end{table*}

\label{sec:sota_comparison}

\subsubsection{Quantitative Comparison}
\label{sec:classwise}
Tables~\ref{tab:classwise_up}--\ref{tab:classwise_hanchuan} compare Compact48
with ten competing classifiers in terms of class-wise accuracy, OA, AA, and
$\kappa$. All results are reported as the mean $\pm$ standard deviation over ten runs, and the highest and second-highest mean values in each row are highlighted in bold and
underlined, respectively. Compact48 consistently achieves the highest OA, AA, and $\kappa$ across all four datasets. In particular, its OA exceeds the strongest competing result by
0.59, 0.35, 0.64, and 1.81 percentage points, respectively. The corresponding
AA advantages are 0.92, 0.45, 0.45, and 3.95 percentage points. These results demonstrate that Compact48 improves not only overall pixel-wise classification accuracy but also average recognition performance across land-cover classes, with particularly pronounced gains on the heterogeneous HanChuan dataset.

The class-wise results further highlight the advantages of Compact48 across
specific land-cover categories. On UP, Compact48 reaches 98.09\% for
Gravel and 98.70\% for Self-Blocking Bricks, improving upon the strongest
competing results for those classes by 2.29 and 1.85 percentage points.
On Houston, the gains are evident in urban transportation categories:
Road and Highway attain 97.37\% and 99.81\%, respectively, representing
improvements of 2.37 and 0.73 percentage points. On Salinas, Compact48
reaches 98.09\% for Vinyard untrained, 2.75 percentage points above the
best competing value. The class-level differences are particularly large
on HanChuan, where Watermelon and Plastic reach 94.40\% and 98.64\%,
exceeding their strongest competitors by 12.71 and 7.37 percentage
points, respectively. These findings demonstrate that the overall performance gains are accompanied by notable improvements in the recognition of specific land-cover categories, particularly on the challenging HanChuan dataset.

\par\vspace{0.15em}

\subsubsection{Qualitative Comparison}
\label{sec:full_qualitative}
Figs.~\ref{fig:app_qual_up}--\ref{fig:app_qual_hanchuan} present the
full-scene classification maps for the competing methods and Compact48.
For each method, the displayed map is selected from the ten independent runs based on its proximity to the mean OA. These visual comparisons provide a spatial perspective on the classification results, complementing the overall and class-wise accuracy comparisons reported in Tables~\ref{tab:classwise_up}--\ref{tab:classwise_hanchuan}.

The UP scene contains large homogeneous areas interspersed with small
objects and narrow structures. As shown in Fig.~\ref{fig:app_qual_up},
several competing methods produce scattered labels within otherwise
coherent regions, whereas Compact48 yields more spatially consistent
predictions while better preserving localized structural details.

Houston contains sparsely distributed labeled areas and thin urban
structures. The comparison in Fig.~\ref{fig:app_qual_houston} is therefore particularly informative within the labeled regions, where differences in predictions are manifested as isolated label substitutions and local inconsistencies.
Compact48 exhibits better visual agreement with the ground truth across
these spatially separated areas.

The Salinas scene consists of large agricultural fields with adjacent
land-cover types. As shown in Fig.~\ref{fig:app_qual_salinas}, some
competing methods exhibit isolated misclassifications within field
interiors. In contrast, Compact48 yields more spatially consistent predictions within these fields while preserving the distinctions between neighboring fields.

HanChuan contains large homogeneous regions, closely spaced cultivated
parcels, and elongated structures. As shown in Fig.~\ref{fig:app_qual_hanchuan},
Compact48 exhibits fewer fragmented predictions in several local regions
and more spatially consistent predictions along narrow structures. These qualitative observations across all four datasets are consistent with the improvements in overall and selected class-wise accuracies reported in Tables~\ref{tab:classwise_up}--\ref{tab:classwise_hanchuan}.

\subsubsection{Computational Cost Comparison}
\label{sec:cross_method_cost}
Table~\ref{tab:cross_method_complexity} compares parameter counts,
FLOPs, full training times, and test inference times of the eleven classifiers
on the four datasets. Compact48 has 75.31--92.07 K parameters
and requires 13.66--31.74 M FLOPs. Its parameter count is consistently lower
than that of SSRN, CVSSN, SSFTT, GAHT, CMT, MambaHSI, and MambaLG,
while its computational cost in terms of FLOPs is substantially lower than that of SSRN, GAHT, and
MambaHSI. Although 3DSS-Mamba and SliMamba have fewer parameters, Compact48 achieves higher OA than both methods across all four datasets (Tables~\ref{tab:classwise_up}--\ref{tab:classwise_hanchuan}).

The measured inference times further reveal variations in computational efficiency. Compact48 achieves inference times of 0.90, 0.46,
2.79, and 11.99~s on UP, Houston, Salinas, and HanChuan, respectively.
Together with the accuracy--compactness comparison in
Fig.~\ref{fig:accuracy_compactness}, these results illustrate the balance between classification accuracy and computational cost achieved by Compact48
relative to the competing methods. The effect of KRGF within the same
architecture is examined in Sec.~\ref{sec:krgf_mechanism}.

\subsection{Effectiveness of Uncertainty-Aware Global--Local Learning}
\label{sec:ablation}

\subsubsection{Complementarity of Global and Local Representations}
We first evaluate whether the global and local encoders provide
complementary information for pixel-wise classification. The Local-only
and Global-only variants retain only their respective encoder branches, whereas
the Dual-view variant integrates both branches. Table~\ref{tab:dual_view_metrics}
shows that Local-only outperforms Global-only on UP and Houston, while Global-only
performs better on Salinas and HanChuan. These differences indicate that
the relative usefulness of center-neighborhood detail and broader
context varies across the evaluated scenes. Dual-view exceeds the
stronger single-branch variant by 2.80, 1.79, 2.21, and 1.08 percentage
points in OA on UP, Houston, Salinas, and HanChuan, respectively.
These results demonstrate the complementarity of global and local representations and motivate their adaptive integration.

\begin{table}[!htbp]
\centering
\caption{Ablation of global and local representations on four HSI datasets
    in terms of OA (\%). The best results are shown in bold}
\label{tab:dual_view_metrics}
\fontsize{7pt}{7.9pt}\selectfont
\setlength{\tabcolsep}{1.4pt}
\renewcommand{\arraystretch}{1.02}
\begin{tabular*}{\columnwidth}{@{\extracolsep{\fill}}cccccc@{}}
\toprule
Local & Global & UP & Houston & Salinas & HanChuan \\
\midrule
\ding{51} & \ding{55} & 96.71$\pm$0.85 & 96.25$\pm$0.37 & 94.84$\pm$1.11 & 96.82$\pm$0.31 \\
\ding{55} & \ding{51} & 95.14$\pm$1.01 & 94.21$\pm$0.77 & 96.23$\pm$0.82 & 97.92$\pm$0.19 \\
\ding{51} & \ding{51} & \textbf{99.51$\pm$0.12} & \textbf{98.04$\pm$0.31} & \textbf{98.44$\pm$0.90} & \textbf{99.00$\pm$0.14} \\
\bottomrule
\end{tabular*}
\end{table}

\subsubsection{Effectiveness of Kalman-Reliable Fusion}
Given the complementary representations, we compare four fusion
strategies: summation (Sum), concatenation (Concat), gated fusion
(Gate), and the proposed KRF. Table~\ref{tab:krf_metrics} shows that
KRF achieves the highest OA on all four datasets. Its improvements
over the strongest alternative are 0.08, 0.22, 0.50, and 0.06
percentage points on UP, Houston, Salinas, and HanChuan, respectively.
These results support the effectiveness of KRF in combining global and local features. To further investigate its fusion mechanism, we examine the contribution of the learned Kalman gain through a controlled comparison.

\begin{table}[!htbp]
\centering
\caption{Comparison of global--local fusion strategies on four HSI datasets
    in terms of OA (\%). The best results are shown in bold}
\label{tab:krf_metrics}
\fontsize{7.1pt}{8.0pt}\selectfont
\setlength{\tabcolsep}{2pt}
\renewcommand{\arraystretch}{1.02}
\begin{tabular*}{\columnwidth}{@{\extracolsep{\fill}}ccccc@{}}
\toprule
Dataset & Sum & Concat & Gate & KRF \\
\midrule
UP & 99.43$\pm$0.21 & 99.33$\pm$0.45 & 99.35$\pm$0.30 & \textbf{99.51$\pm$0.12} \\
Houston & 97.59$\pm$0.59 & 97.64$\pm$0.54 & 97.82$\pm$0.44 & \textbf{98.04$\pm$0.31} \\
Salinas & 97.85$\pm$1.12 & 96.95$\pm$2.58 & 97.94$\pm$1.12 & \textbf{98.44$\pm$0.90} \\
HanChuan & 98.72$\pm$0.17 & 98.82$\pm$0.17 & 98.94$\pm$0.11 & \textbf{99.00$\pm$0.14} \\
\bottomrule
\end{tabular*}
\end{table}

To isolate the contribution of uncertainty-dependent weighting, we
further compare the learned KRF with a fixed-gain variant using
$\mathbf{K}=0.5$ in the posterior feature update. All other fusion components and data partitions remain unchanged, and the learned-KRF results are obtained from the original Dense64 models. Houston and Salinas provide complementary cases because their
stronger single-branch variants are Local-only and Global-only,
respectively (Table~\ref{tab:dual_view_metrics}). As shown in
Table~\ref{tab:fixed_gain_metrics}, the learned KRF improves mean OA
from 97.92\% to 98.04\% on Houston and from 97.94\% to 98.44\%
on Salinas. These controlled results indicate that adapting the Kalman gain according to learned relative uncertainties yields better classification performance than using a fixed global--local weighting scheme.

\begin{table}[!htbp]
\centering
\caption{Comparison of fixed and learned Kalman gains in Dense64
    in terms of OA (\%; mean $\pm$ standard deviation over ten runs)}
\label{tab:fixed_gain_metrics}
\footnotesize
\setlength{\tabcolsep}{4pt}
\renewcommand{\arraystretch}{1.08}
\begin{tabular*}{\columnwidth}{@{\extracolsep{\fill}}ccc@{}}
\toprule
Dataset & Fixed $K=0.5$ & KRF \\
\midrule
Houston & 97.92$\pm$0.35 & \textbf{98.04$\pm$0.31} \\
Salinas & 97.94$\pm$1.27 & \textbf{98.44$\pm$0.90} \\
\bottomrule
\end{tabular*}
\end{table}

\subsubsection{Relationship Between Kalman Gain and Branch-Level Prediction Correctness}
We further investigate whether the learned Kalman gain varies
consistently with the relative predictive performance of
the global and local representations. The trained Dense64
encoders are held fixed. Specifically, separate linear probes are trained on pooled features from the global and local encoders using the training split, and calibrated using the validation split. These probes characterize branch-level prediction outcomes
without modifying the learned fusion mechanism.

We focus on test pixels where the global and local probes differ in prediction correctness and compare the channel-wise and spatially
averaged Kalman gain under two conditions: global correct/local
incorrect ($G^{\checkmark}L^{\times}$) and global incorrect/local
correct ($G^{\times}L^{\checkmark}$). According to
\eqref{eq:krf_update}, a larger gain assigns greater weight
to the local representation in the posterior update.
Therefore, comparing the conditional mean gains reveals
whether KRF tends to favor local information when its
corresponding branch prediction is correct. This analysis
further characterizes the relationship between learned
fusion weights and branch-level predictive behavior.

\begin{table}[!htbp]
\centering
\caption{Kalman gain under different branch-prediction outcomes
    (mean $\pm$ standard deviation over ten runs). The conditional averages
    correspond to the global and local linear-probe predictions; $\Delta K$
    denotes the second mean minus the first}
\label{tab:gain_branch_correctness}
\fontsize{7.2pt}{8.2pt}\selectfont
\setlength{\tabcolsep}{2.3pt}
\renewcommand{\arraystretch}{1.08}
\begin{tabular*}{\columnwidth}{@{\extracolsep{\fill}}cccc@{}}
\toprule
Dataset & $G^{\checkmark}L^{\times}$ & $G^{\times}L^{\checkmark}$ & $\Delta K$ \\
\midrule
Houston & 0.5140$\pm$0.0188 & 0.5681$\pm$0.0148 & +0.0541$\pm$0.0150 \\
Salinas & 0.5234$\pm$0.0181 & 0.5567$\pm$0.0187 & +0.0333$\pm$0.0122 \\
\bottomrule
\end{tabular*}
\end{table}

As shown in Table~\ref{tab:gain_branch_correctness}, the mean Kalman gain is
higher when the local probe is correct and the global probe is
incorrect, with differences of 0.0541 on Houston and 0.0333
on Salinas. This consistent trend suggests that the learned gain is associated with
the relative predictive quality of the two branches. Together
with the fixed-gain comparison, these results provide complementary evidence, in terms of both classification performance and fusion behavior, supporting the uncertainty-
dependent update employed by KRF.

\subsection{Sensitivity to Global and Local View Sizes}
\label{sec:patchsize}
Table~\ref{tab:patchsize_sensitivity} compares paired global/local
view sizes of $15/5$, $21/7$, and $25/9$ while keeping the hidden width fixed. Within each dataset, the parameter count is unchanged as
spatial support increases, whereas FLOPs increase with the spatial extent of the input patches. On UP, extending the views from $15/5$ to $21/7$ improves mean OA
from 99.02\% to 99.51\%. A further increase to $25/9$ yields a
similar OA (99.50\%) but raises FLOPs from 21.60 to 31.96 M.
Houston likewise reaches its highest mean OA with $21/7$ views.
For Salinas and HanChuan, the largest evaluated views yield the
highest mean OA (98.44\% and 99.00\%, respectively), with higher
computational costs.

The different trends show that the appropriate spatial support
is scene dependent: broader context may add useful information,
but its extra cost does not guarantee a classification gain.
According to the validation-based configuration used throughout
the experiments, the global/local sizes are set to $21/7$ for
UP and Houston and $25/9$ for Salinas and HanChuan.

\begin{table}[!htbp]
\centering
\caption{Sensitivity to global and local view sizes. The highest OA, AA,
    and $\kappa\times100$ for each dataset are shown in bold}
\label{tab:patchsize_sensitivity}
\begingroup
\scriptsize
\setlength{\tabcolsep}{1.1pt}
\renewcommand{\arraystretch}{1.12}
\begin{tabular*}{\columnwidth}{@{\extracolsep{\fill}}cccccc@{}}
\toprule
Dataset & G/L & OA (\%) & AA (\%) & $\kappa\times100$ & FLOPs (M) \\
\midrule
\multirow{3}{*}{UP}
 & 15/5 & 99.02$\pm$0.29 & 98.83$\pm$0.38 & 98.71$\pm$0.38 & 11.24 \\
 & 21/7 & \textbf{99.51$\pm$0.12} & \textbf{99.32$\pm$0.17} & \textbf{99.35$\pm$0.16} & 21.60 \\
 & 25/9 & 99.50$\pm$0.10 & 99.26$\pm$0.30 & 99.34$\pm$0.13 & 31.96 \\
\midrule
\multirow{3}{*}{Houston}
 & 15/5 & 97.89$\pm$0.38 & 98.00$\pm$0.56 & 97.72$\pm$0.41 & 12.55 \\
 & 21/7 & \textbf{98.04$\pm$0.31} & \textbf{98.12$\pm$0.47} & \textbf{97.88$\pm$0.33} & 24.18 \\
 & 25/9 & 97.98$\pm$0.44 & 98.05$\pm$0.63 & 97.81$\pm$0.48 & 35.66 \\
\midrule
\multirow{3}{*}{Salinas}
 & 15/5 & 96.90$\pm$1.17 & 98.55$\pm$0.55 & 96.55$\pm$1.30 & 15.11 \\
 & 21/7 & 97.60$\pm$0.99 & 98.86$\pm$0.49 & 97.33$\pm$1.10 & 29.19 \\
 & 25/9 & \textbf{98.44$\pm$0.90} & \textbf{99.20$\pm$0.47} & \textbf{98.27$\pm$1.00} & 42.89 \\
\midrule
\multirow{3}{*}{HanChuan}
 & 15/5 & 98.05$\pm$0.22 & 96.70$\pm$0.55 & 97.72$\pm$0.26 & 16.71 \\
 & 21/7 & 98.71$\pm$0.19 & 97.80$\pm$0.49 & 98.50$\pm$0.22 & 32.33 \\
 & 25/9 & \textbf{99.00$\pm$0.14} & \textbf{98.12$\pm$0.43} & \textbf{98.83$\pm$0.16} & 47.41 \\
\bottomrule
\end{tabular*}
\endgroup
\vspace{-0.5ex}
\end{table}

\begin{figure*}[!t]
\centering
\includegraphics[width=\textwidth]{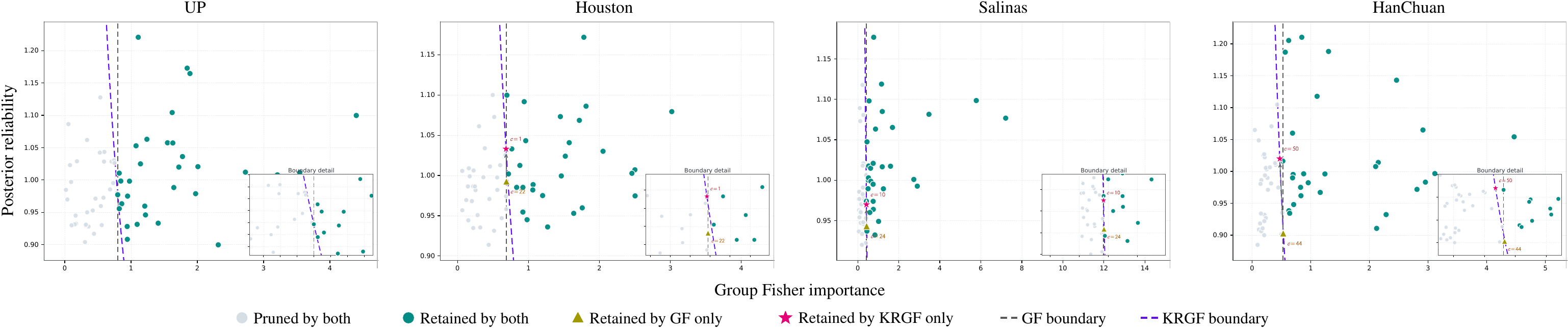}
\caption{Coupled-group selections under $64\rightarrow32$ compression on
    four HSI datasets. The horizontal and vertical axes show Group Fisher importance
    and posterior reliability, respectively. Markers identify groups retained or
    pruned by both criteria or selected exclusively by one criterion; dashed lines
    indicate the displayed selection boundaries. Insets magnify the boundary regions.
    Each panel shows a run whose KRGF OA is closest to its ten-run mean.}
\label{fig:fisher_reliability_selection}
\end{figure*}

\begin{figure*}[!t]
\centering
\includegraphics[width=\textwidth]{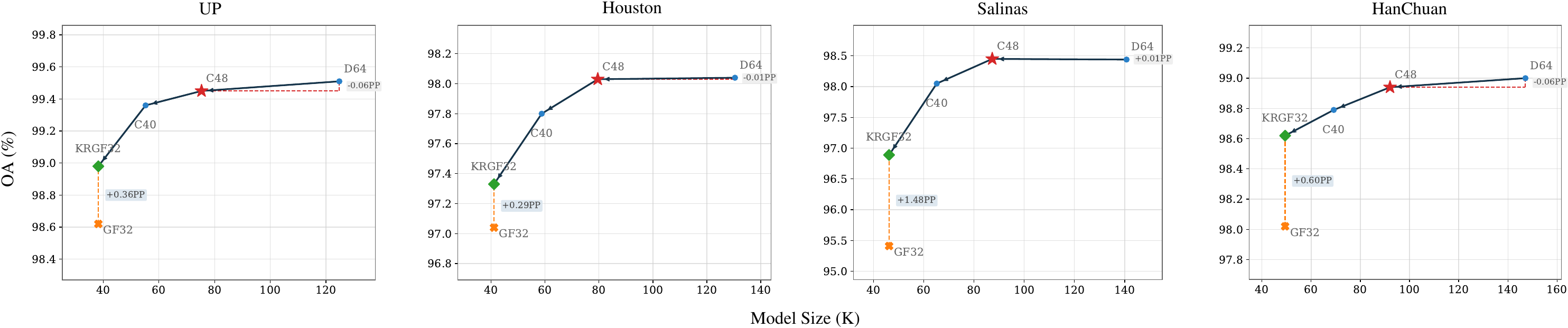}
\caption{Accuracy--model size trade-offs under progressive KRGF compression
    on UP, Houston, Salinas, and HanChuan. The axes denote parameter count (K)
    and OA (\%). D64 represents Dense64, while C48 and C40 represent Compact48
    and Compact40. KRGF32 and GF32 denote width-32 models selected using KRGF and
    Group Fisher importance alone, respectively.}
\label{fig:compression_efficiency}
\end{figure*}

\subsection{Reliability-Guided Pruning and Compression Efficiency}
\label{sec:krgf_mechanism}

\subsubsection{Effectiveness of Reliability-Weighted Group Selection}
We first examine individual and combined contributions of Group Fisher importance and posterior reliability to the KRGF score $Q_c=\mathcal{I}_c\rho_c$ under the common
$64\!\rightarrow\!32$ compression setting. Table~\ref{tab:krgf_metrics}
compares Group Fisher importance $\mathcal{I}_c$, normalized
posterior reliability $\rho_c$, and their product. Compared with
Fisher-only selection, the combined criterion improves OA by
0.36, 0.29, 1.48, and 0.60 percentage points on UP, Houston,
Salinas, and HanChuan, respectively. It also exceeds the
stronger individual criterion on each dataset, by 0.36, 0.29,
1.48, and 0.40 percentage points, respectively. These results
indicate that task sensitivity and estimated posterior reliability
provide complementary information for selecting coupled
structures under strong compression.

\begin{table}[!htbp]
\centering
\caption{Comparison of coupled-group selection criteria under
    $64\!\rightarrow\!32$ compression in terms of OA (\%).
    The best results are shown in bold}
\label{tab:krgf_metrics}
\fontsize{7pt}{7.9pt}\selectfont
\setlength{\tabcolsep}{1.0pt}
\renewcommand{\arraystretch}{1.04}
\begin{tabular*}{\columnwidth}{@{\extracolsep{\fill}}cccc@{}}
\toprule
Dataset & \shortstack{Group Fisher\\($\mathcal{I}_c$)} & \shortstack{Posterior Reliability\\($\rho_c$)} & KRGF ($Q_c$) \\
\midrule
UP & 98.62$\pm$0.71 & 97.36$\pm$0.99 & \textbf{98.98$\pm$0.32} \\
Houston & 97.04$\pm$0.56 & 96.66$\pm$0.61 & \textbf{97.33$\pm$0.32} \\
Salinas & 95.41$\pm$2.47 & 92.18$\pm$4.98 & \textbf{96.89$\pm$0.86} \\
HanChuan & 98.02$\pm$0.31 & 98.22$\pm$0.28 & \textbf{98.62$\pm$0.20} \\
\bottomrule
\end{tabular*}
\end{table}

\begin{table}[!htbp]
\begingroup
\centering
\caption{Classification performance and computational efficiency of
    KGLP-Net under progressive compression on four HSI datasets. The final
    row reports the average change of Compact48 relative to Dense64}
\label{tab:compression_efficiency_full}

\fontsize{6.7pt}{7.5pt}\selectfont
\setlength{\tabcolsep}{1.0pt}
\renewcommand{\arraystretch}{1.02}
\begin{tabular*}{\columnwidth}{@{\extracolsep{\fill}}cccccc@{}}
\toprule
Dataset & Model & OA (\%) & Params (K) & FLOPs (M) & Time (s) \\
\midrule
\multirow[c]{4}{*}{UP}
& Dense64 & 99.51$\pm$0.12 & 124.80 & 21.60 & 2.02 \\
& \textbf{Compact48} & 99.45$\pm$0.12 & 75.31 & 13.66 & 0.90 \\
& Compact40 & 99.36$\pm$0.20 & 55.22 & 10.33 & 0.76 \\
& Compact32 & 98.98$\pm$0.32 & 38.24 & 7.41 & 0.63 \\
\midrule
\multirow[c]{4}{*}{Houston}
& Dense64 & 98.04$\pm$0.31 & 130.44 & 24.18 & 0.70 \\
& \textbf{Compact48} & 98.03$\pm$0.21 & 79.54 & 15.59 & 0.46 \\
& Compact40 & 97.80$\pm$0.36 & 58.75 & 11.93 & 0.39 \\
& Compact32 & 97.33$\pm$0.32 & 41.06 & 8.70 & 0.33 \\
\midrule
\multirow[c]{4}{*}{Salinas}
& Dense64 & 98.44$\pm$0.90 & 140.74 & 42.89 & 3.78 \\
& \textbf{Compact48} & 98.45$\pm$0.83 & 87.27 & 28.35 & 2.79 \\
& Compact40 & 98.05$\pm$0.85 & 65.19 & 22.03 & 2.66 \\
& Compact32 & 96.89$\pm$0.86 & 46.21 & 16.35 & 2.41 \\
\midrule
\multirow[c]{4}{*}{HanChuan}
& Dense64 & 99.00$\pm$0.14 & 147.14 & 47.41 & 18.32 \\
& \textbf{Compact48} & 98.94$\pm$0.14 & 92.07 & 31.74 & 11.99 \\
& Compact40 & 98.79$\pm$0.16 & 69.19 & 24.86 & 12.22 \\
& Compact32 & 98.62$\pm$0.20 & 49.41 & 18.61 & 12.04 \\
\midrule
\multicolumn{2}{c}{\textbf{Avg. $\Delta$ (48 vs. 64)}}
& \textbf{$-0.03$ pp} & \textbf{$-38.5\%$} & \textbf{$-34.8\%$} & \textbf{$-37.6\%$} \\
\bottomrule
\end{tabular*}
\endgroup

\end{table}

\subsubsection{Effect of Posterior Reliability on Coupled-Group Selection}
We further investigate how reliability weighting affects coupled-group selection. Fig.~\ref{fig:fisher_reliability_selection} presents
the Top-32 coupled groups selected by Fisher-only and KRGF for
representative runs. The rankings differ primarily around the
retention boundary, where the Fisher scores of candidate groups
are relatively close. For Houston, group $c=22$
is selected by Fisher-only, whereas $c=1$ is selected by KRGF.
Similar selection differences are observed on Salinas and HanChuan, while both criteria retain identical groups in the representative UP run. These observations illustrate how posterior reliability influences coupled-group selection, providing structural evidence that complements the classification results in Table~\ref{tab:krgf_metrics}.

\subsubsection{Accuracy--Efficiency Trade-Off Under Progressive Compression}
Finally, we evaluate the accuracy--efficiency trade-off of Dense64 and the progressively reconstructed
Compact48, Compact40, and Compact32 models.
Table~\ref{tab:compression_efficiency_full} reports their OA,
parameter counts, FLOPs, and inference times, while
Fig.~\ref{fig:compression_efficiency} shows the accuracy--model-size
trajectories. Relative to Dense64, Compact48 exhibits OA changes of
$-0.06$, $-0.01$, $+0.01$, and $-0.06$ percentage points on
UP, Houston, Salinas, and HanChuan, respectively. Across these
datasets, it reduces the parameter count by 38.5\%, FLOPs by
34.8\%, and inference time by 37.6\% on average, with only a
0.03-percentage-point decrease in mean OA. These results identify Compact48 as a favorable accuracy--efficiency operating point
among the evaluated widths.

Further compression to widths 40 and 32 reduces parameter counts and FLOPs, generally at the cost of larger OA decreases. Inference times
also decrease overall, although runtime does not change
monotonically with FLOPs on every scene, reflecting the
execution characteristics of the compact models. At width 32, KRGF consistently outperforms Fisher-only selection across the four datasets, further supporting the effectiveness of reliability-weighted group selection under strong compression. Fig.~\ref{fig:accuracy_compactness}
places Compact48 alongside the ten competing classifiers in
terms of OA, model size, and FLOPs.

\section{Conclusion}
\label{sec:conclusion}

This article presented KGLP-Net, a unified framework that integrates uncertainty-aware
global--local fusion with reliability-guided structured pruning for
HSI classification. Its asymmetric Mamba--CNN encoders capture complementary global contextual information and local spectral--spatial features. KRF estimates the relative feature-space uncertainties of these representations and performs a Kalman-inspired posterior update, while KRGF incorporates the resulting posterior reliability into Group Fisher importance for coupled-group selection. Thus, the uncertainty information obtained during feature fusion also guides subsequent structured model compression.

Experiments on four public HSI datasets validate the benefits of dual-view
learning, KRF, and reliability-weighted group selection. Compact48
achieves the highest OA among ten competing classifiers. Compared with Dense64, it reduces the parameter count, FLOPs, and inference time by
38.5\%, 34.8\%, and 37.6\% on average, respectively, while incurring only a 0.03-percentage-point decrease in mean OA. These results highlight the favorable accuracy--efficiency trade-off achieved by combining uncertainty-aware representation learning with reliability-guided structured pruning. Future work will extend this principle toward
input-adaptive inference, allocating computation according to
estimated prediction risks and contextual requirements of HSI samples.

\end{document}